\documentclass{article}
\usepackage{xcolor}

\usepackage[final]{corl_2019} 
\usepackage{graphicx}
\usepackage{fullpage,enumitem,amsmath,amssymb, mathtools, amsthm,wrapfig}
\usepackage{array}
\usepackage{booktabs}
\usepackage{makecell}
\usepackage{multirow}
\usepackage{lscape}
\usepackage{rotating}

\usepackage{tabularx}
\usepackage{booktabs}
\usepackage{multicol}
\usepackage{letltxmacro}
\usepackage{siunitx}
\usepackage{placeins}
\usepackage{hyperref}
\usepackage{float}

\newcommand{\RLsq}{\texorpdfstring{RL${}^2$}{RL2}}

\newcolumntype{L}[1]{>{\raggedright\arraybackslash}p{#1}}
\newcolumntype{C}[1]{>{\centering\arraybackslash}p{#1}}
\newcolumntype{R}[1]{>{\raggedleft\arraybackslash}p{#1}}
\LetLtxMacro\oldttfamily\ttfamily
\DeclareRobustCommand{\ttfamily}{\oldttfamily\csname ttsize\endcsname}
\newcommand{\setttsize}[1]{\def\ttsize{#1}}%

\usepackage{titlesec}
\titlespacing\section{0pt}{3pt plus 4pt minus 2pt}{0pt plus 2pt minus 2pt}
\titlespacing\subsection{0pt}{3pt plus 4pt minus 2pt}{0pt plus 2pt minus 2pt}
\titlespacing\subsubsection{0pt}{3pt plus 4pt minus 2pt}{0pt plus 2pt minus 2pt}

\usepackage{fleqn, tabularx}
\usepackage{multirow}
\usepackage[export]{adjustbox}

\makeatletter

\providecommand{\todo}[1]{{\color{red} [TODO: #1]}}

\newcommand\blfootnote[1]{%
  \begingroup
  \renewcommand\thefootnote{}\footnote{#1}%
  \addtocounter{footnote}{-1}%
  \endgroup
}

\title{ Meta-World:  A Benchmark and Evaluation for \\ Multi-Task and Meta Reinforcement Learning
}

\author{
  Tianhe Yu$^{*1}$, Deirdre Quillen$^{*2}$, Zhanpeng He$^{*3}$, Ryan Julian$^{*4}$, Avnish Narayan$^{*4}$, \\ \textbf{Hayden Shively}$^{4}$, \textbf{Adithya Bellathur}$^{4}$, \\\textbf{Karol Hausman}$^5$, \textbf{Chelsea Finn}$^1$, \textbf{Sergey Levine}$^2$\\
  Stanford University$^1$, UC Berkeley$^2$, Columbia University$^3$,\\ University of Southern California$^4$, Robotics at Google$^5$  \\  
}

\begin{document}
\maketitle


\begin{abstract}
\blfootnote{$*$ denotes equal contribution}
Meta-reinforcement learning algorithms can enable robots to acquire new skills much more quickly, by leveraging prior experience to learn how to learn. However, much of the current research on meta-reinforcement learning focuses on task distributions that are very narrow. For example, a commonly used meta-reinforcement learning benchmark uses different running velocities for a simulated robot as different tasks. When policies are meta-trained on such narrow task distributions, they cannot possibly generalize to more quickly acquire entirely new tasks. Therefore, if the aim of these methods is enable faster acquisition of entirely new behaviors, we must evaluate them on task distributions that are sufficiently broad to enable generalization to new behaviors. In this paper, we propose an open-source simulated benchmark for meta-reinforcement learning and multi-task learning consisting of 50 distinct robotic manipulation tasks. Our aim is to make it possible to develop algorithms that generalize to accelerate the acquisition of entirely new, held-out tasks. We evaluate 7 state-of-the-art meta-reinforcement learning and multi-task learning algorithms on these tasks. 
Surprisingly, while each task and its variations (e.g., with different object positions) can be learned with reasonable success, these algorithms struggle to learn with multiple tasks at the same time, even with as few as ten distinct training tasks. Our analysis and open-source environments pave the way for future research in multi-task learning and meta-learning that can enable meaningful generalization, thereby unlocking the full potential of these methods.\footnote{Videos of the benchmark tasks are on the anonymous project page: \url{meta-world.github.io}.} 

\blfootnote{Our open-sourced code for the benchmark is available at: \url{https://github.com/rlworkgroup/metaworld}.} 
\blfootnote{All of the open-sourced baselines and launchers for
our experiments can be found at \url{https://github.com/rlworkgroup/garage}.}

\blfootnote{This manuscript is an update on a manuscript that appeared at the 3rd Conference on Robot Learning (CoRL 2019), Osaka, Japan.}
\end{abstract}



\section{Introduction}
\vspace{-0.2cm}
While reinforcement learning (RL) has achieved some success in domains such as assembly~\cite{DBLP:journals/corr/LevineFDA15}, ping pong~\cite{mulling2013learning}, in-hand manipulation~\cite{andrychowicz2018learning}, and hockey~\cite{chebotar2017combining}, state-of-the-art methods require substantially more experience than humans to acquire only one narrowly-defined skill. If we want robots to be broadly useful in realistic environments, we instead need algorithms that can learn a wide variety of skills reliably and efficiently. Fortunately, in most specific domains, such as robotic manipulation or locomotion, many individual tasks share common structure that can be reused to acquire related tasks more efficiently. 
For example, most robotic manipulation tasks involve grasping or moving objects in the workspace. However, while current methods can learn to individual skills like screwing on a bottle cap \cite{DBLP:journals/corr/LevineFDA15} and hanging a mug \cite{DBLP:journals/corr/abs-1903-06684}, we need algorithms that can efficiently learn shared structure across many related tasks, and use that structure to learn new skills quickly, such as screwing a jar lid or hanging a bag.
Recent advances in machine learning have provided unparalleled  generalization capabilities in domains such as images~\cite{Krizhevsky:2017:ICD:3098997.3065386} and speech~\cite{DBLP:journals/corr/abs-1810-04805}, suggesting that this should be possible; however, we have yet to see such generalization to diverse tasks in reinforcement learning settings.

Recent works in meta-learning and multi-task reinforcement learning have shown promise for addressing this gap. 
Multi-task RL methods aim to learn a single policy that can solve multiple tasks more efficiently than learning the tasks individually, while meta-learning methods train on many tasks, and optimize for fast adaptation to a new task.
While these methods have made progress, the development of both classes of approaches has been limited by the lack of established benchmarks and evaluation protocols that reflect realistic use cases. 
On one hand, multi-task RL methods have largely been evaluated on disjoint and overly diverse tasks such as the Atari suite  \cite{DBLP:journals/corr/abs-1809-04474}, where there is little efficiency to be gained by learning across games~\cite{parisotto2015actor}.
On the other hand, meta-RL methods have been evaluated on very narrow task distributions. For example, one popular evaluation of meta-learning involves choosing different running directions for simulated legged robots~\cite{finn2017model}, which then enables fast adaptation to new directions. While these are technically distinct tasks, they are a far cry from the promise of a meta-learned model that can adapt to any new task within some domain.
In order to study the capabilities of current multi-task and meta-reinforcement learning methods and make it feasible to design new algorithms that actually generalize and adapt quickly on meaningfully distinct tasks, we need evaluation protocols and task suites that are broad enough to enable this sort of generalization, while containing sufficient shared structure for generalization to be possible.

The key contributions of this work are a suite of $50$ diverse simulated manipulation tasks and an extensive empirical evaluation of how previous methods perform on sets of such distinct tasks.
We contend that multi-task and meta reinforcement learning methods that aim to efficiently learn many tasks and quickly generalize to new tasks should be evaluated on distributions of tasks that are diverse and exhibit shared structure. 
To this end, we present a benchmark of simulated manipulation tasks with everyday objects, all of which are contained in a shared, table-top environment with a simulated Sawyer arm.
By providing a large set of distinct tasks that share common environment and control structure, we believe that this benchmark will allow researchers to test the generalization capabilities of the current multi-task and meta RL methods, and help to identify new research avenues to improve the current approaches.
Our empirical evaluation of existing methods on this benchmark reveals that, despite some impressive progress in multi-task and meta-reinforcement learning over the past few years, current methods are generally not able to learn diverse task sets, much less generalize successfully to entirely new tasks. We provide an evaluation protocol with evaluation modes of varying difficulty, and observe that current methods show varying amounts of success on these modes
This opens the door for future developments in multi-task and meta reinforcement learning: instead of focusing on further increasing performance on current narrow task suites, we believe that it is essential for future work in these areas to focus on increasing the capabilities of algorithms to handle highly diverse task sets.

By doing so, we can enable meaningful generalization across many tasks and achieve the full potential of meta-learning as a means of incorporating past experience to make it possible for robots to acquire new skills as quickly as people can.

\begin{figure}[t]
    \centering
    \vspace{-0.4cm}
    \includegraphics[width=\columnwidth]{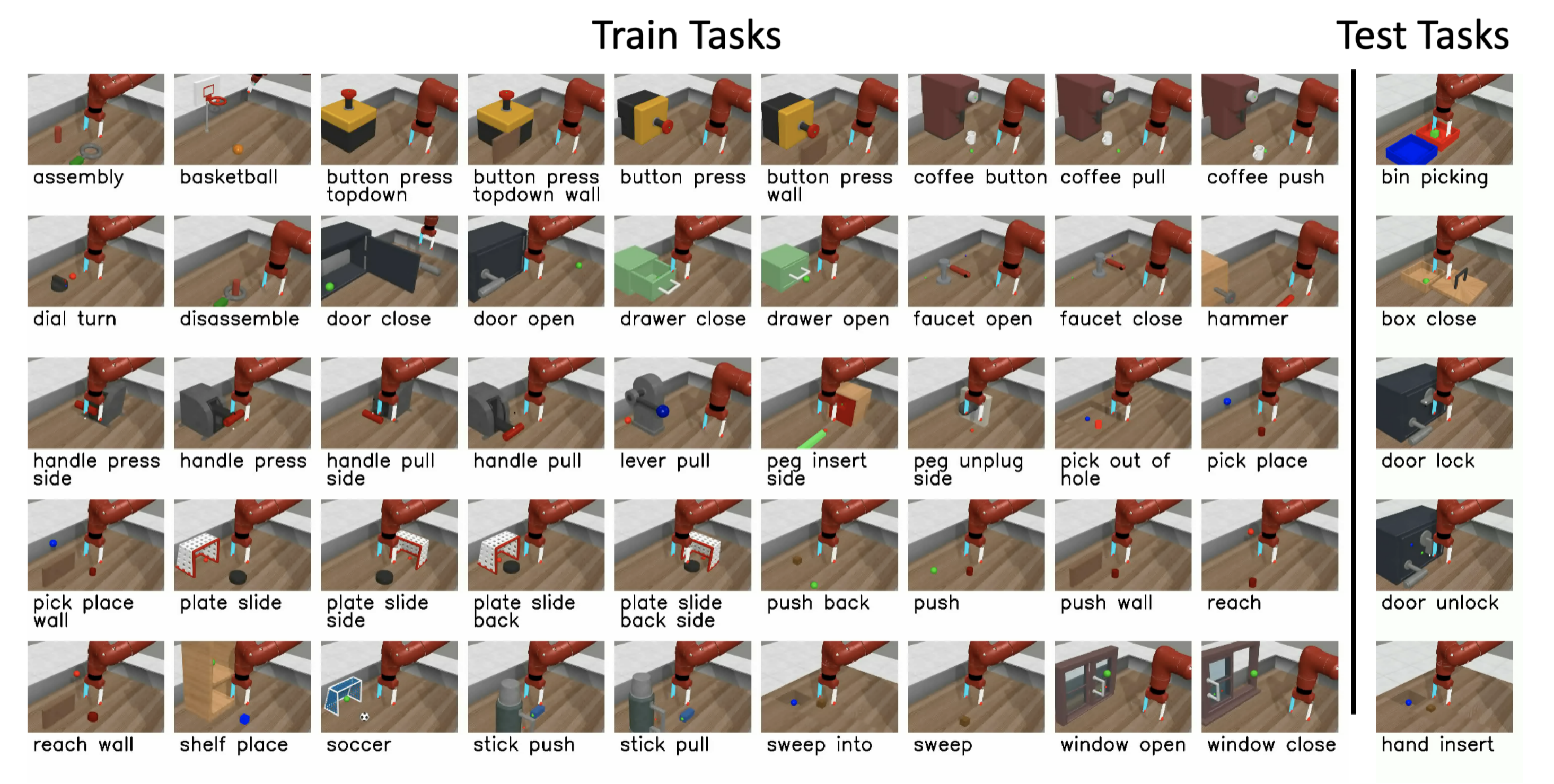}
    \vspace{-0.35cm}
    \caption{Meta-World contains 50 manipulation tasks, designed to be diverse yet carry shared structure that can be leveraged for efficient multi-task RL and transfer to new tasks via meta-RL. In the most difficult evaluation, the method must use experience from 45 training tasks (left) to quickly learn distinctly new test tasks (right). A larger view of the environments can be found on the next page.
    }
    \label{fig:ml45_teaser}
    \vspace{-0.5cm}
\end{figure}


\section{Related Work}
\vspace{-0.2cm}




Previous works that have proposed benchmarks for reinforcement learning have largely focused on single task learning settings~\cite{brockman2016openai,cobbe2018quantifying,tassa2018deepmind}.
One popular benchmark used to study multi-task learning is the Arcade Learning Environment, a suite of dozens of Atari 2600 games~\cite{DBLP:journals/corr/abs-1709-06009}.
While having a tremendous impact on the multi-task reinforcement learning research community~\cite{parisotto2015actor,rusu2015policy,DBLP:journals/corr/abs-1809-04474,espeholt2018impala,sharma2017online}, the Atari games included in the benchmark have significant differences in visual appearance, controls, and objectives, making it challenging to acquire any efficiency gains through shared learning.
In fact, many prior multi-task learning methods have observed substantial negative transfer between the Atari games~\cite{parisotto2015actor,rusu2015policy}.
In contrast, we would like to study a case where positive transfer between the different tasks should be possible. 
We therefore propose a set of related yet diverse tasks that share the same robot, action space, and workspace.

Meta-reinforcement learning methods have been evaluated on a number of different problems, including maze navigation~\cite{DBLP:journals/corr/DuanSCBSA16,wang1611learning,mishra2017simple},
continuous control domains with parametric variation across tasks~\cite{finn2017model,rothfuss2018promp,rakelly2019efficient,fernando2018meta}, bandit problems~\cite{wang1611learning,DBLP:journals/corr/DuanSCBSA16,mishra2017simple,ritter2018been}, levels of an arcade game~\cite{nichol2018gotta}, and locomotion tasks with varying dynamics~\cite{nagabandi2018learning,saemundsson2018meta}.
Complementary to these evaluations, we aim to develop a testbed of tasks and an evaluation protocol that are reflective of the challenges in applying meta-learning to robotic manipulation problems, including both parameteric and non-parametric variation in tasks.


There is a long history of robotics benchmarks~\cite{calli2015benchmarking,DBLP:journals/corr/abs-1911-07246,james2019rlbench}, datasets~\cite{lenz2015deep,finn2016unsupervised,yu2016more,chebotar2016bigs,gupta2018robot,mandlekar2018roboturk,sharma2018multiple}, competitions~\cite{correll2016analysis} and standardized object sets~\cite{calli2015ycb,choi2009list} that have played an important role in robotics research.
Similarly, there exists a number of robotics simulation benchmarks including visual navigation~\cite{savva2019habitat,kolve2017ai2,brodeur2017home,savva2017minos,xia2018gibson}, autonomous driving~\cite{dosovitskiy2017carla,wymann2000torcs,richter2017playing}, grasping~\cite{kappler2015leveraging,kasper2012kit,goldfeder2008columbia}, single-task manipulation~\cite{corl2018surreal}, among others.
In this work, our aim is to continue this trend and provide a large suite of tasks that will allow researchers to study multi-task learning, meta-learning, and transfer in general. Further, unlike these prior simulation benchmarks, we particularly focus on providing a suite of many diverse manipulation tasks and a protocol for multi-task and meta RL evaluation.

\section{The Multi-Task and Meta-RL Problem Statements}
\label{sec:problem}
\vspace{-0.2cm}

Our proposed benchmark is aimed at making it possible to study generalization in meta-RL and multi-task RL. In this section, we define the meta-RL and multi-task RL problem statements, and describe some of the challenges associated with task distributions in these settings.


\newcommand{\task}{\mathcal{T}}

We use the formalism of Markov decision processes (MDPs), where each task $\task$ corresponds to a different finite horizon MDP, represented by a tuple $(S, A, P, R, H, \gamma)$, where $s \in S$ correspond to states,
$a \in A$ correspond to the available actions, $P(s_{t+1}|s_t, a_t)$ represents the stochastic transition dynamics,
$R(s, a)$ is a reward function, $H$ is the horizon and $\gamma$ is the discount factor. In standard reinforcement learning, the goal is to learn a policy $\pi(a|s)$ that maximizes the expected return, which is the sum of (discounted) rewards over all time.
In multi-task and meta-RL settings, we assume a distribution of tasks $p(\task)$. Different tasks may vary in any aspect of the Markov decision process, though efficiency gains in adaptation to new tasks are only possible if the tasks share some common structure. For example, as we describe in the next section, the tasks in our proposed benchmark have the same action space and horizon, and structurally similar rewards and state spaces.\footnote{In practice, the policy must be able to read in the state for each of the tasks, which typically requires them to at least have the same dimensionality. In our benchmarks, some tasks have different numbers of objects, but the state dimensionality is always the same, meaning that some state coordinates are unused for some tasks.}

\textbf{Multi-task RL problem statement.} 
The goal of multi-task RL is to learn a single, task-conditioned policy $\pi(a|s, z)$, where $z$ indicates an encoding of the task ID. This policy should maximize the average expected return across all tasks from the task distribution $p(\task)$, given by $\mathbb{E}_{\task \sim p(\task)} [\mathbb{E}_{\pi}[\sum_{t=0}^T \gamma^t R_t(s_t, a_t)]]$.
The information about the task can be provided to the policy in various ways, e.g. using a one-hot task identification encoding $z$ that is passed in addition to the current state.
There is no separate test set of tasks, and multi-task RL algorithms are typically evaluated on their average performance over the \emph{training} tasks.

\textbf{Meta-RL problem statement.} 
Meta-reinforcement learning aims to leverage the set of training task to learn a policy $\pi(a|s)$ that can quickly adapt to new test tasks that were not seen during training, where both training and test tasks are assumed to be drawn from the same task distribution $p(\task)$. Typically, the training tasks are referred to as the \emph{meta-training} set, to distinguish from the adaptation (training) phase performed on the (meta-) test tasks.
During meta-training, the learning algorithm has access to $M$ tasks $\{\task_i\}_{i=1}^M$ that are drawn from the task distribution $p(\task)$.
At meta-test time, a new task $\task_j \sim p(\task)$ is sampled that was not seen during meta-training, and the meta-trained policy must quickly adapt to this task to achieve the highest return with a small number of samples.
A key premise in meta-RL is that a sufficiently powerful meta-RL method can meta-learn a model that effectively implements a highly efficient reinforcement learning procedure, which can then solve entirely new tasks very quickly -- much more quickly than a conventional reinforcement learning algorithm learning from scratch. However, in order for this to happen, the meta-training distribution $p(\task)$ must be sufficiently broad to encompass these new tasks. Unfortunately, most prior work in meta-RL evaluates on very narrow task distributions, with only one or two dimensions of parametric variation, such as the running direction for a simulated robot~\cite{finn2017model,rothfuss2018promp,rakelly2019efficient,fernando2018meta}.

\newcommand{\1}{\mathbb{I}}

\section{Meta-World}
\vspace{-0.2cm}


If we want meta-RL methods to generalize effectively to entirely new tasks, we must meta-train on broad task distributions that are representative of the range of tasks that a particular agent might need to solve in the future. 
To this end, we propose a new multi-task and meta-RL benchmark, which we call Meta-World. In this section, we motivate the design decisions behind the Meta-World tasks, discuss
the range of tasks, describe the representation of the actions, observations, and rewards, and present a set of evaluation protocols of varying difficulty for both meta-RL and multi-task RL.

\subsection{The Space of Manipulation Tasks: Parametric and Non-Parametric Variability}
\label{sec:parametric}

\begin{wrapfigure}{r}{0.5\textwidth}
    \centering
    \includegraphics[width=0.5\columnwidth]{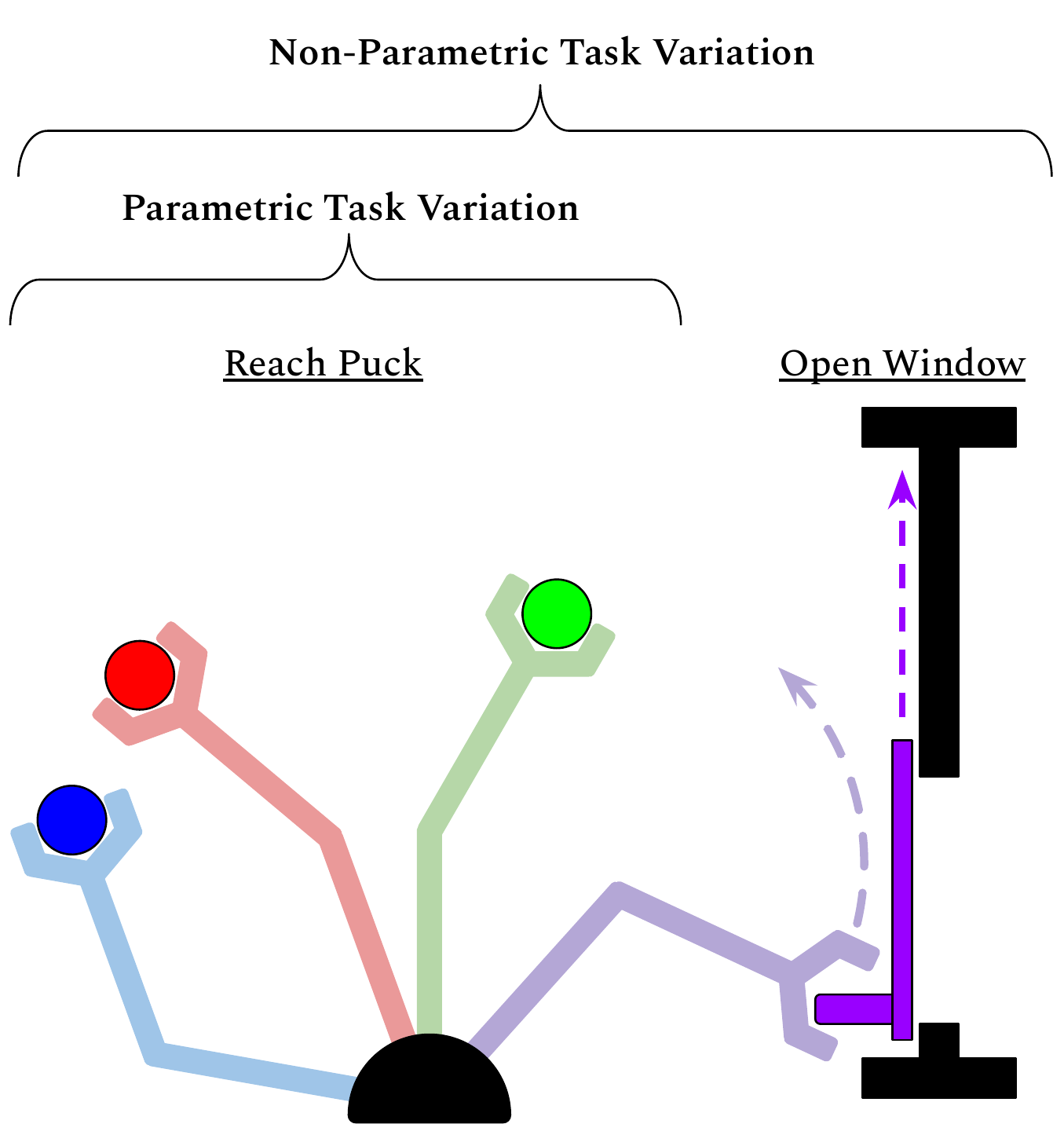}
    \caption{\footnotesize Parametric/non-parametric variation: all ``reach puck'' tasks (left) can be parameterized by the puck position, while the difference between ``reach puck'' and ``open window'' (right) is non-parametric.}
    \vspace{-0.5cm}
    \label{fig:variation-catoon}
\end{wrapfigure}


A task, $\task$, in Meta-World is defined as the tuple \textit{(reward\; function, \;initial \;object \;position, \;target \;position)}
Meta-learning makes two critical assumptions: first, that the meta-training and meta-test tasks are drawn from the same distribution, $p(\task)$, and second, that the task distribution $p(\task)$ exhibits shared structure that can be utilized for efficient adaptation to new tasks. 
If $p(\task)$ is defined as a family of variations within a particular control task, as in prior work~\cite{finn2017model,rakelly2019efficient}, then it is unreasonable to hope for generalization to entirely new control tasks. For example, an agent has little hope of being able to quickly learn to open a door, without having ever experienced doors before, if it has only been trained on a set of meta-training tasks that are homogeneous and narrow. 
Thus, to enable meta-RL methods to adapt to entirely new tasks, we propose a much larger suite of tasks consisting of $50$ qualitatively-distinct manipulation tasks, where continuous parameter variation cannot be used to describe the differences between tasks.

With such non-parametric variation, however, there is the danger that tasks will not exhibit enough shared structure, or will lack the task overlap needed for the method to avoid memorizing each of the tasks. 
Motivated by this challenge, we design each task to include parametric variation in object and goal positions, as illustrated in Figure~\ref{fig:variation-catoon}.
Introducing this parametric variability not only creates a substantially larger (infinite) variety of tasks, but also makes it substantially more practical to expect that a meta-trained model will generalize to acquire entirely new tasks more quickly, since varying the positions provides for wider coverage of the space of possible manipulation tasks. Without parametric variation, the model could for example memorize that any object at a particular location is a door, while any object at another location is a drawer. 
If the locations are not fixed, this kind of memorization is much less likely, and the model is forced to generalize more broadly. 
With enough tasks and variation within tasks, pairs of qualitatively-distinct tasks are more likely to overlap, serving as a catalyst for generalization.
For example, closing a drawer and pushing a block can appear as nearly the same task for some initial and goal positions of each object.

Note that this kind of parametric variation, which we introduce \emph{for each task}, essentially represents the entirety of the task distribution for previous meta-RL evaluations~\cite{finn2017model, rakelly2019efficient}, which test on single tasks (e.g., running towards a goal) with parametric variability (e.g., variation in the goal position). Our full task distribution is therefore substantially broader, since it includes this parametric variability \emph{for each of the $50$ tasks}.

To provide shared structure, the $50$ environments require the same robotic arm to interact with different objects, with different shapes, joints, and connectivity. The tasks themselves require the robot to execute a combination of reaching, pushing, and grasping, depending on the task.
By recombining these basic behavioral building blocks with a variety of objects with different shapes and articulation properties, we can create a wide range of manipulation tasks. For example, the \textbf{open door} task
involves pushing or grasping an object with a revolute joint, while the \textbf{open drawer} task requires pushing or grasping an object with a sliding joint.
More complex tasks require a combination of these building blocks, which must be executed in the right order.
We visualize all of the tasks in Meta-World in Figure~\ref{fig:ml45_teaser}, and include a description of all tasks in Appendix~\ref{app:tasks}.

All of the tasks are implemented in the MuJoCo physics engine~\cite{todorov2012mujoco}, which enables fast simulation of physical contact. 
To make the interface simple and accessible, we base our suite on the Multiworld interface~\cite{nair2018visual} and the OpenAI Gym environment interfaces~\cite{brockman2016openai}, making additions and adaptations of the suite relatively easy for researchers already familiar with Gym.

\subsection{Actions, Observations, and Rewards}


In order to represent policies for multiple tasks with one model, the observation and action spaces must contain significant shared structure across tasks. All of our tasks are performed by a simulated Sawyer robot. The action space is a 2-tuple consisting of the change in 3D space of the end-effector followed by a normalized torque that the gripper fingers should apply. The actions in this space range between $-1$ and $1$.
For all tasks, the robot must either manipulate one object with a variable goal position, or manipulate two objects with a fixed goal position. The observation space is represented as a 6-tuple of the 3D Cartesian positions of the end-effector, a normalized measurement of how open the gripper is, the 3D position of the first object, the quaternion of the first object, the 3D position of the second object, the quaternion of the second object, all of the previous measurements in the environment, and finally the 3D position of the goal. If there is no second object or the goal is not meant to be included in the observation, then the quantities corresponding to them are zeroed out. The observation space is always $39$ dimensional.



Designing reward functions for Meta-World requires two major considerations. First, to guarantee that our tasks are within the reach of current single-task reinforcement learning algorithms, which is a prerequisite for evaluating multi-task and meta-RL algorithms, 
we design well-shaped reward functions for each task that make each of the tasks at least individually solvable.
More importantly, the reward functions must exhibit shared structure across tasks. Critically, even if the reward function admits the same optimal policy for multiple tasks, varying reward scales or structures can make the tasks appear completely distinct for the learning algorithm, masking their shared structure and leading to preferences for tasks with high-magnitude rewards~\cite{DBLP:journals/corr/abs-1809-04474}. 
Accordingly, we adopt a structured, multi-component
reward function for all tasks, which leads to effective policy learning for each of the task components. For instance, in a task that involves a combination of reaching, grasping, and placing an object, let $o \in \mathbb{R}^3$ be the object position, where $o = (o_x, o_y, o_z)$, $h \in \mathbb{R}^3$ be the position of the robot's gripper, $z_\text{target} \in \mathbb{R}$ be the target height of lifting the object, and $g \in \mathbb{R}^3$ be goal position. With the above definition, the multi-component reward function $R$ is the combination of a reaching reward, a grasping reward, and a placing reward or subsets thereof for simpler tasks that only involve reaching and/or pushing.
With this design, the reward functions across all tasks have a similar magnitude that ranges between $0$ and $10$, where $10$ always corresponds to the reward-function being solved, and conform to similar structure, as desired. The full form of the reward function and a list of all task rewards is provided in Appendix~\ref{app:rewardfns}.

\subsection{Evaluation Protocol}
\begin{figure}
    \centering
    \includegraphics[width=\columnwidth]{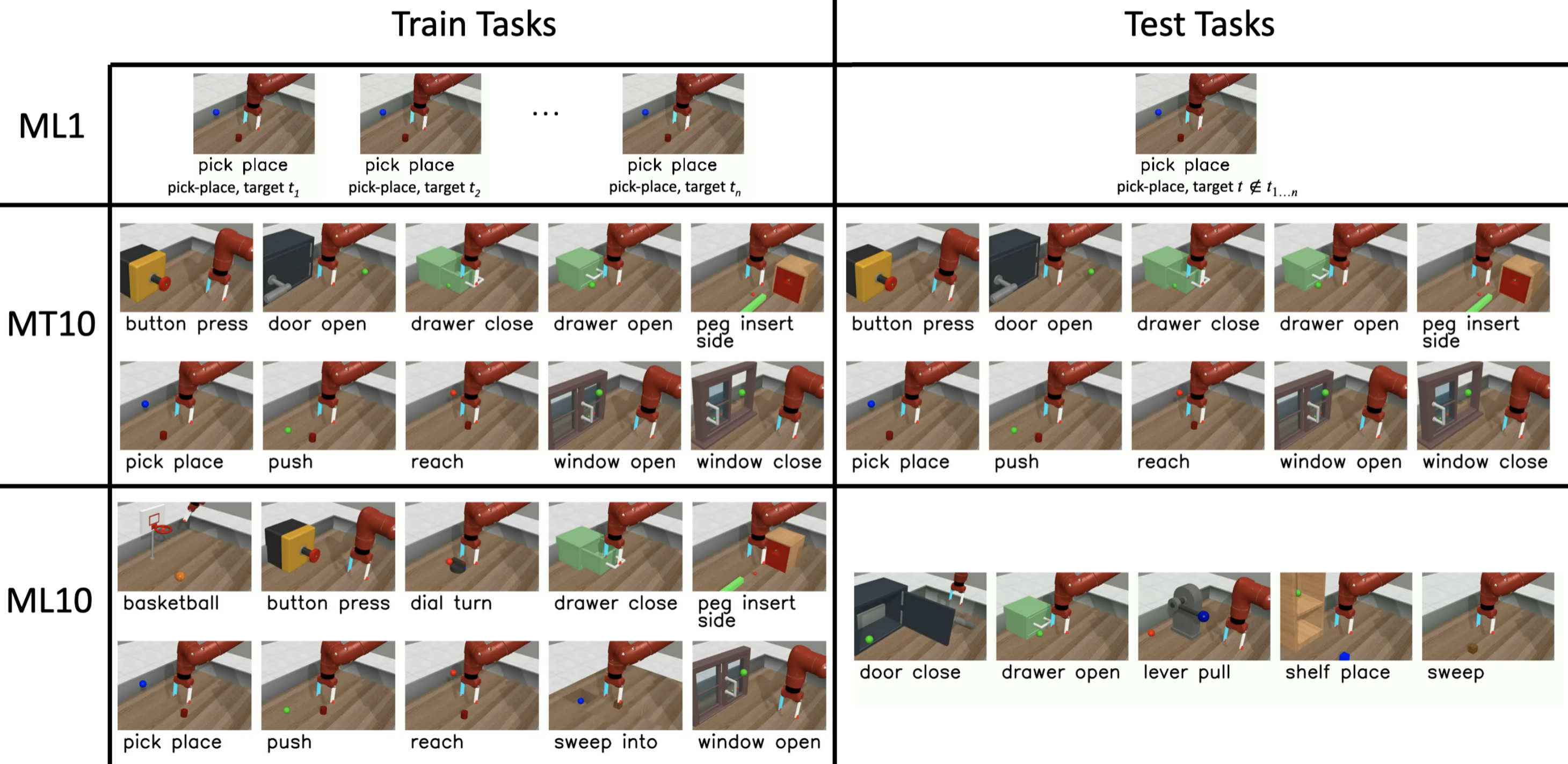}
    \caption{\footnotesize Visualization of three of our multi-task and meta-learning evaluation protocols, ranging from within task adaptation in ML1, to multi-task training across 10 distinct task families in MT10, to adapting to new tasks in ML10. Our most challenging evaluation mode ML45 is shown in Figure~\ref{fig:ml45_teaser}.
    }
    \label{fig:evaluation}
\end{figure}
With the goal of providing a challenging benchmark to facilitate progress in multi-task RL and meta-RL, we design an evaluation protocol with varying levels of difficulty, ranging from the level of current goal-centric meta-RL benchmarks to a setting where methods must learn distinctly new, challenging manipulation tasks based on diverse experience across 45 tasks. We hence divide our evaluation into five categories, which we describe next. We then detail our evaluation criteria.


\textbf{Meta-Learning 1 (ML1): Few-shot adaptation to goal variation within one task.} The simplest evaluation aims to verify that previous meta-RL algorithms can adapt to new object or goal configurations on only one type of task. ML1 uses single Meta-World Tasks, with the meta-training ``tasks'' corresponding to $50$ random initial object and goal positions, and meta-testing on $50$ held-out positions. This resembles the evaluations in prior works~\cite{finn2017model, rakelly2019efficient}. We evaluate algorithms on three individual tasks from Meta-World: reaching, pushing, and pick and place, where the variation is over reaching position or goal object position. The goal positions are not provided in the observation, forcing meta-RL algorithms to adapt to the goal through trial-and-error.

\textbf{Multi-Task 1 (MT1): Learning one multi-task policy that generalizes to 50 tasks belonging to the same environment}. This evaluation aims to verify how well multi-task algorithms can learn across a large related task distribution. MT1 uses single Meta-World environments, with the training ``tasks'' corresponding to $50$ random initial object and goal positions. The goal positions are provided in the observation and are a fixed set, as to focus on the ability of algorithms in acquiring a distinct skill across multiple goals, rather than generalization and robustness.

\textbf{Multi-Task 10, Multi-Task 50 (MT10, MT50): Learning one multi-task policy that generalizes to 50 tasks belonging to 10 and 50 training environments, for a total of 500, and 2,500 training tasks.} A first step towards adapting quickly to distinctly new tasks is the ability to train a single policy that can solve multiple distinct training tasks. The multi-task evaluation in Meta-World tests the ability to learn multiple tasks at once, without accounting for generalization to new tasks. The MT10 evaluation uses 10 environments: reach, push, pick and place, open door, open drawer, close drawer, press button top-down, insert peg side, open window, and open box. The larger MT50 evaluation uses all $50$ Meta-World environments. In our experiments, the algorithm is typically provided with a one-hot vector indicating the current task. The positions of objects and goal positions are fixed in all tasks in this evaluation, so as to focus on acquiring the distinct skills, rather than generalization and robustness.


\textbf{Meta-Learning 10, Meta-Learning 45 (ML10, ML45): Few-shot adaptation to new test tasks with 10 and 50 meta-training tasks.}  With the objective to test generalization to new tasks, we hold out 5 tasks and meta-train policies on 10 and 45 tasks.
We randomize object and goals positions 
and intentionally select training tasks with structural similarity to the test tasks. Task IDs are not provided as input, requiring a meta-RL algorithm to identify the tasks from experience.


\textbf{Success metrics.} 
Since values of reward are not directly indicative how successful a policy is, we define an interpretable success metric for each task, which will be used as the evaluation criterion for all of the above evaluation settings.
Since all of our tasks involve manipulating one or more objects into a goal configuration, this success metric is typically based on the distance between the task-relevant object and its final goal pose, i.e. $\|o - g\|_2 < \epsilon$, where $\epsilon$ is a small distance threshold such as $5$ cm. For the complete list of success metrics and thresholds for each task, see Appendix~\ref{tbl:task_metrics}.




\section{Experimental Results and Analysis}
\vspace{-0.2cm}
\label{sec:result}

The first, most basic goal of our experiments is to verify that each of the $50$ presented tasks are indeed solveable by existing single-task reinforcement learning algorithms. We provide this verification in Appendix~\ref{app:singletask}. Beyond verifying the individual tasks, the goals of our experiments are to study the following questions: (1) can existing state-of-the-art meta-learning algorithms quickly learn qualitatively new tasks when meta-trained on a sufficiently broad, yet structured task distribution, and (2) how do different multi-task and meta-learning algorithms compare in this setting? 
To answer these questions, we evaluate various multi-task and meta-learning algorithms on the Meta-World benchmark. We include the training curves of all evaluations in Figure~\ref{fig:learningcurves} in the Appendix~\ref{app:curves}. Videos of the tasks and evaluations, along with all source code, are on the project webpage\footnote{Videos are on the project webpage,  at \url{meta-world.github.io}
}.

In the multi-task evaluation, we evaluate the following RL algorithms:
\textbf{multi-task proximal policy optimization (PPO)} \cite{schulman2017proximal}: a policy gradient algorithm adapted to the multi-task setting by providing the one-hot task ID as input, \textbf{multi-task trust region policy optimization (TRPO)} \cite{schulman2015trust}: an on-policy policy gradient algorithm adapted to the multi-task setting using the one-hot task ID as input, \textbf{multi-task soft actor-critic (SAC)}~\cite{haarnoja2018soft}: an off-policy actor-critic algorithm adapted to the multi-task setting using the one-hot task ID as input, and an on-policy version of \textbf{task embeddings (TE)}~\cite{hausman2018learning}: a multi-task reinforcement learning algorithm that parameterizes the learned policies via shared skill embedding space.
For the meta-RL evaluation, we study three algorithms:
\textbf{RL$^2$} \cite{DBLP:journals/corr/DuanSCBSA16,wang1611learning}: an on-policy meta-RL algorithm that corresponds to training a GRU network with hidden states maintained across episodes within a task and trained with PPO, \textbf{model-agnostic meta-learning (MAML)} \cite{finn2017model, rothfuss2018promp}: an on-policy gradient-based meta-RL algorithm that embeds policy gradient steps into the meta-optimization, and is trained with PPO, and \textbf{probabilistic embeddings for actor-critic RL (PEARL)} \cite{rakelly2019efficient}: an off-policy actor-critic meta-RL algorithm, which learns to encode experience into a probabilistic embedding of the task that is fed to the actor and the critic. We use the baselines in the Garage \cite{garage} reinforcement learning library, which we developed for benchmarking Meta-World.

\begin{figure}[H]
    \centering
    \includegraphics[width=\columnwidth]{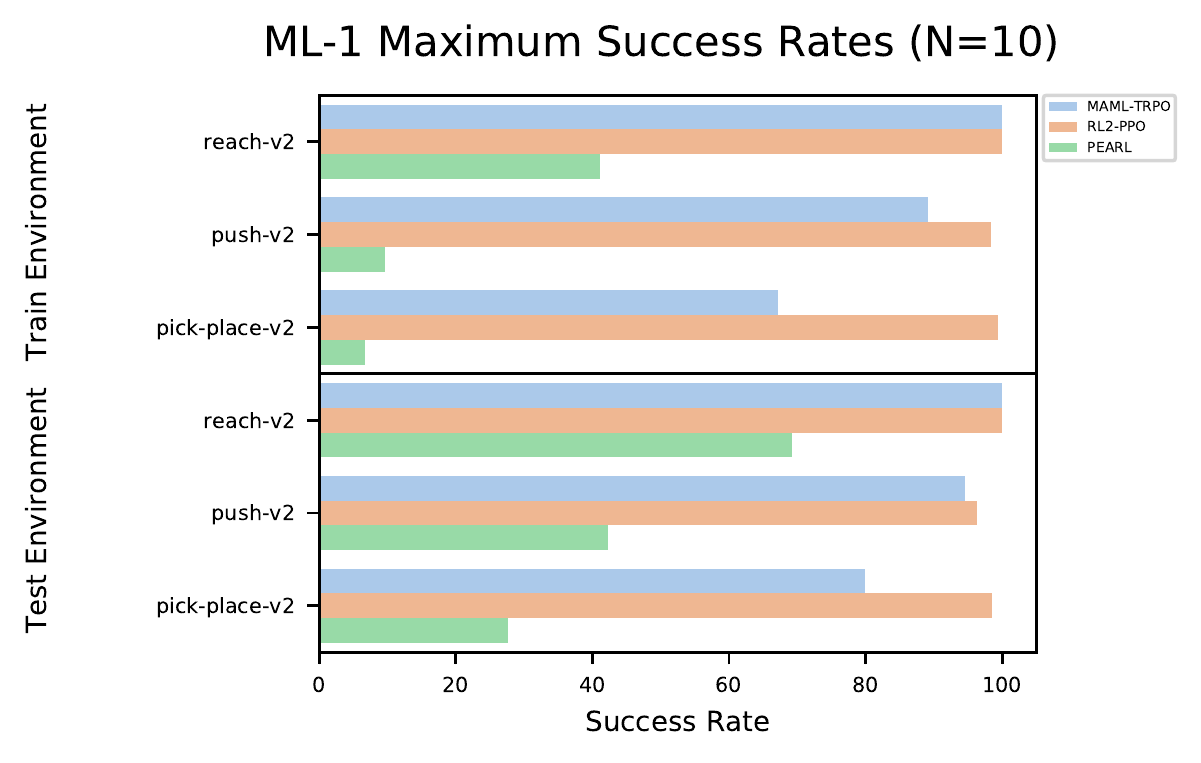}
    \vspace{-0.5cm}
    \caption{\footnotesize{Comparison on our simplest meta-RL evaluation, ML1 on 10 seeds. RL$^2$ shows the strongest performance in generalization. Pearl shows the weakest performance, though this could be attributed to difficulty in training its task encoder}
    }
    \label{fig:ml1}
\end{figure}

\begin{figure}[H]
    
    \includegraphics[width=\columnwidth]{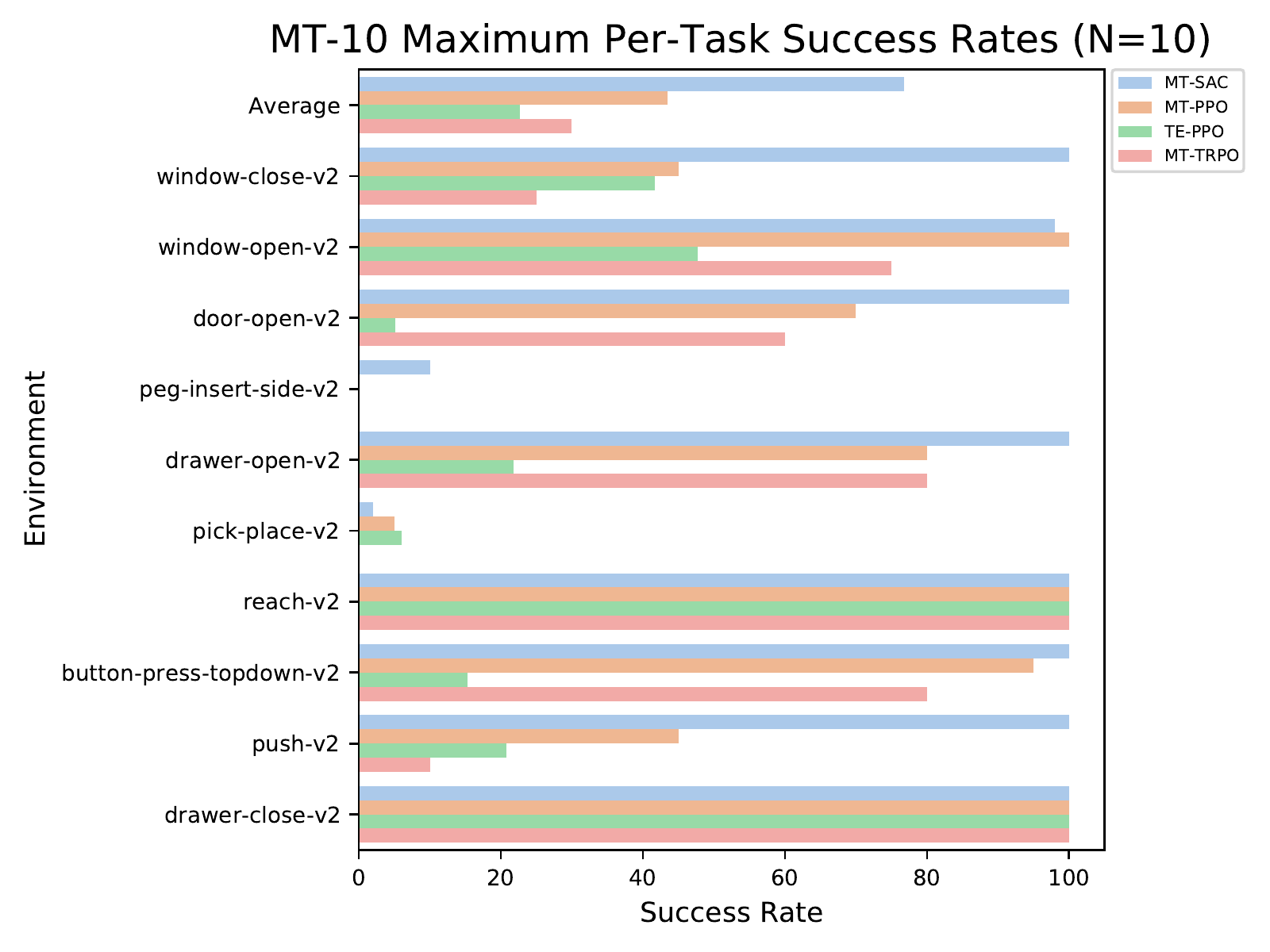}
    \vspace{-1cm}
    \caption{Performance of the tested MTRL algorithms on 10 seeds. MT-SAC performs the best on MT-10, exhibiting the greatest sample efficiency and performance. For detailed plots of these algorithm's learning curves, see appendix \ref{app:curves}.}
    \label{fig:mt10}
\end{figure}

We show results of the simplest meta-learning evaluation mode, ML1, in Figure~\ref{fig:ml1}. We find that there is room for improvement even in this very simple setting. Next, we look at results of multi-task learning across distinct tasks, starting with MT10 in Figure~\ref{fig:mt10} and in Table~\ref{tab:final_results}.
\\
We find that multi-task SAC is able to the learn the MT10 task suite well, achieving around 68\% success rate averaged across tasks, while multi-task PPO and TRPO are only able to achieve around a 30\% success rate. However, as we scale to 50 distict tasks with MT50, we find that MT-SAC and MT-PPO only achieve around a 35-38\% success rate, indicating that there is significant room for improvement in these methods

\begin{figure}[H]
    \includegraphics[width=\columnwidth]{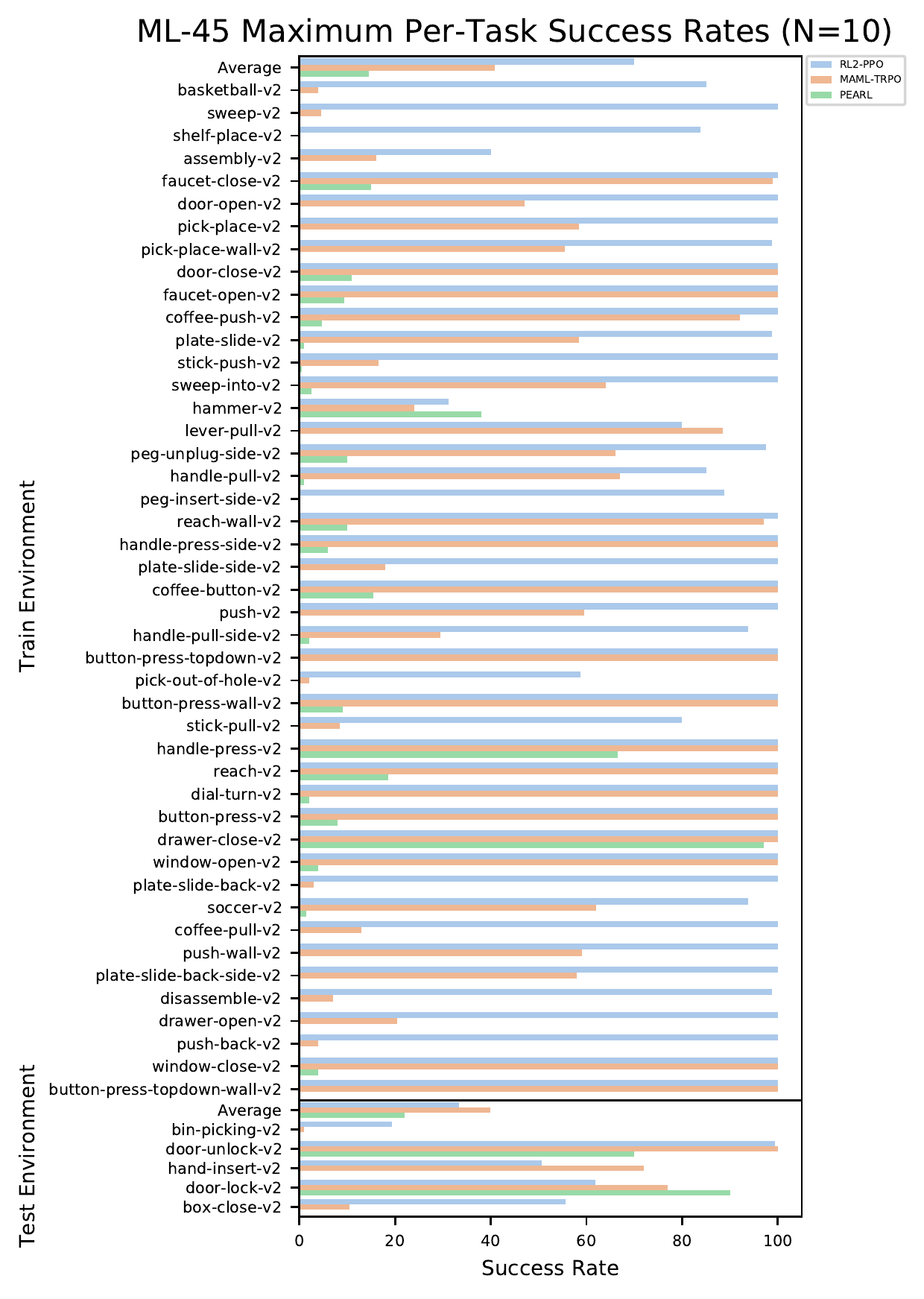}
    \caption{\footnotesize Average of maximum success rate for ML-45. Note that, even on the challenging ML-45 benchmark, current methods already exhibit some degree of generalization, but meta-training performance leaves considerable room for improvement, suggesting that future work could attain better performance on these benchmarks. Though PEARL has week training performance, it has comparable performance on test tasks. RL$^2$ has the highest  We also show the max average success rates for all benchmarks in Table \ref{tab:final_results}.}
    \label{fig:ml45}
\end{figure}

Finally, we study the ML10 and ML45 meta-learning benchmarks, which require learning the meta-training tasks and generalizing to new meta-test tasks with small amounts of experience. From Figure~\ref{fig:ml45} and Table~\ref{tab:final_results}, we find that the prior meta-RL methods, MAML and RL$^2$ reach 35\% and 31\% success on ML10 test tasks, while PEARL achieves only 13\% on ML10.
On ML45, MAML and RL$^2$ solve around 39.9\% and 33.3\% of the meta-test tasks. Note that, on both ML10 and ML45, the meta-training performance of all methods also has considerable room for improvement, suggesting that optimization challenges are generally more severe in the meta-learning setting. The fact that some methods nonetheless exhibit meaningful generalization suggests that the ML10 and ML45 benchmarks are solvable, but challenging for current methods, leaving considerable room for improvement in future work.
\vspace{3cm}
\clearpage
\begin{figure}[H]  
    \includegraphics[width=\columnwidth]{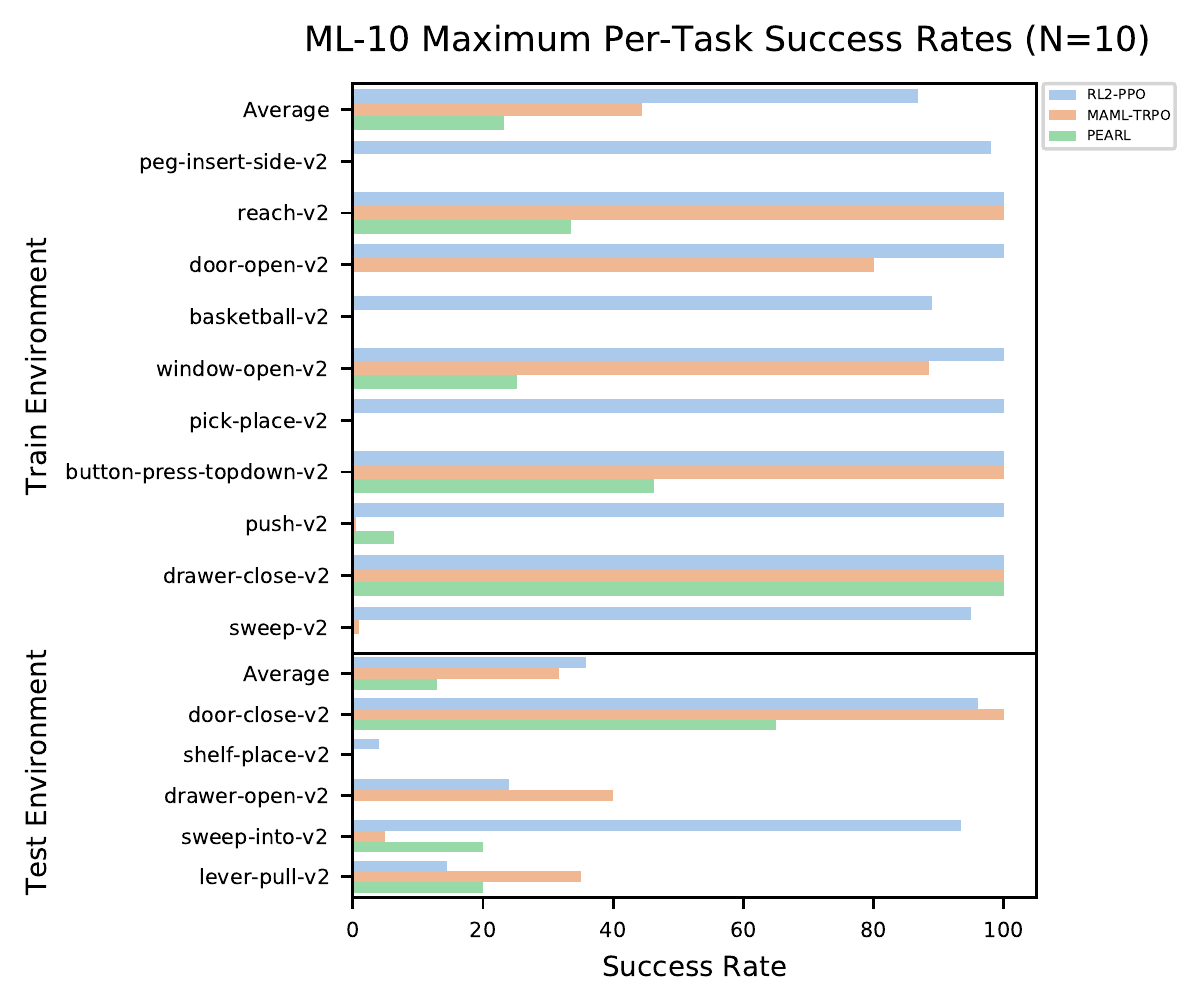}
    \caption{Performance of the tested meta-RL algorithms on 10 seeds. RL$^2$ shows the highest performance on the training tasks (86.9\%), however its ability to generalize is not that much greater than MAML (35.8\% for RL$^2$ and 31.6\% for MAML).}
    \label{fig:ml10}
\end{figure}
\begin{figure}[H]
    \vspace{-1cm}
    \includegraphics[width=\columnwidth]{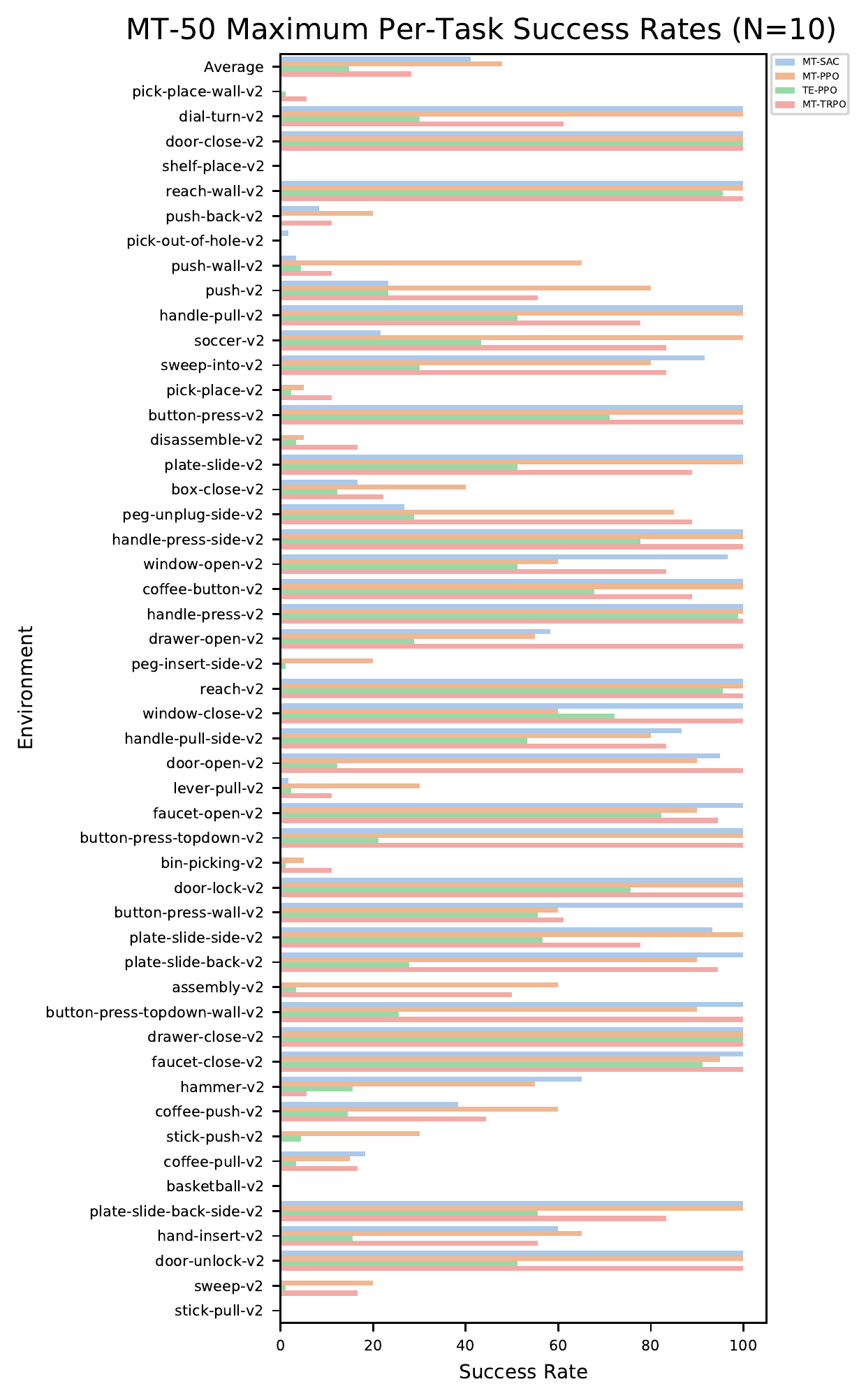}
    \vspace{-0.75cm}
    \caption{Performance of the tested MTRL algorithms on 10 seeds. In MT-10, MT-SAC showed the highest performance, however its performance does not scale to MT-50, the more difficult benchmark. MT-PPO exhibits the better performance in this benchmark.}
    \label{fig:mt50}
\end{figure}

\begin{table*}[h]
  \centering
  \scriptsize
  \def\arraystretch{0.9}
  \setlength{\tabcolsep}{1em}
  \begin{tabularx}{0.43\linewidth}{ccll*{9}{c}}
  \toprule
 \multicolumn{1}{c}{Methods} & \multicolumn{1}{c}{MT10} & \multicolumn{1}{c}{MT50}\\
\midrule
  Multi-task PPO&   30.5\% & \textbf{35.4}\%\\
  Multi-task TRPO&   31.3\% & 21.0\%\\
  Task embeddings&   20.9\% & 11.8\%\\
 Multi-task SAC&   \textbf{68.3}\% & \textbf{38.5}\%\\
    \bottomrule
    \end{tabularx}
    \begin{tabularx}{0.5\linewidth}{ccll*{9}{c}}
  \toprule
 \multicolumn{1}{c}{\multirow{2}[4]{*}{Methods}} & \multicolumn{2}{c}{ML10} & \multicolumn{2}{c}{ML45}\\
    \cmidrule(lr){2-3} \cmidrule(lr){4-5} 
     & \multicolumn{1}{c}{meta-train}  & \multicolumn{1}{c}{meta-test}  & \multicolumn{1}{c}{meta-train} & \multicolumn{1}{c}{meta-test}\\
  \midrule
  MAML&   44.4\%  &31.6\% & 40.7\%  &\textbf{39.9\%}\\
  RL$^2$&   \textbf{86.9\%}  & \textbf{35.8}\% & \textbf{70\%}  &33.3\%\\
  PEARL&   23.2\%  & 13\% & 14.5\%  & 22\%\\
    \bottomrule
    \end{tabularx}
     \caption{\footnotesize The average maximum success rate over all tasks for MT10, MT50, ML10, and ML45 on 10 seeds. The best performance in each benchmark is bolden. For MT10 and MT50, we show the average training success rate of multi-task SAC and multi-task PPO respectively outperform other methods. For ML10 and ML45, we show the meta-train and meta-test success rates. RL$^2$ achieves best meta-train performance in ML10 and ML45, while MAML and RL$2$ get the best generalization performance in ML10 and ML45 meta-test tasks respectively.
     }
    \label{tab:final_results}
\end{table*}

\section{Conclusion and Directions for Future Work}
\label{sec:conclusion}
We proposed an open-source benchmark for meta-reinforcement learning and multi-task learning, which consists of a large number of simulated robotic manipulation tasks.

Unlike previous evaluation benchmarks in meta-RL, our benchmark specifically emphasizes generalization to distinctly new tasks, not just in terms of parametric variation in goals, but completely new objects and interaction scenarios. 

While meta-RL can in principle make it feasible for agents to acquire new skills more quickly by leveraging past experience, previous evaluation benchmarks utilize very narrow task distributions, making it difficult to understand the degree to which meta-RL actually enables this kind of generalization. The aim of our benchmark is to make it possible to develop new meta-RL algorithms that actually exhibit this sort of generalization.
Our experiments show that current meta-RL methods in fact cannot yet generalize effectively to entirely new tasks and do not even learn the meta-training tasks effectively when meta-trained across multiple distinct tasks. This suggests a number of directions for future work, which we describe below.

\textbf{Future directions for algorithm design.}
The main conclusion from our experimental evaluation with our proposed benchmark is that current meta-RL algorithms generally struggle in settings where the meta-training tasks are highly diverse. This issue mirrors the challenges observed in multi-task RL, which is also challenging with our task suite, and has been observed to require considerable additional algorithmic development to attain good results in prior work~\cite{parisotto2015actor,rusu2015policy,espeholt2018impala}.
A number of recent works have studied algorithmic improvements in the area of multi-task reinforcement learning, as well as potential explanations for the difficulty of RL in the multi-task setting~\cite{DBLP:journals/corr/abs-1809-04474,schaul2019ray}.
Incorporating some of these methods into meta-RL, as well as developing new techniques to enable meta-RL algorithms to train on broader task distributions, would be a promising direction for future work to enable meta-RL methods to generalize effectively across diverse tasks, and our proposed benchmark suite can provide future algorithms development with a useful gauge of progress towards the eventual goal of broad task generalization.

\textbf{Future extensions of the benchmark.} While the presented benchmark is significantly broader and more challenging than existing evaluations of meta-reinforcement learning algorithms, there are a number of extensions to the benchmark that would continue to improve and expand upon its applicability to realistic robotics tasks. First, in many situations, the poses of objects are not directly accessible to a robot in the real world.
Hence, one interesting and important direction for future work is to consider image observations and sparse rewards. Sparse rewards can be derived already using the success metrics, while support for image rendering is already supported by the code. However, for meta-learning algorithms, special care needs to be taken to ensure that the task cannot be inferred directly from the image, else meta-learning algorithms will memorize the training tasks rather than learning to adapt.
Another natural extension would be to consider including a breadth of compositional long-horizon tasks, where there exist combinatorial numbers of tasks. Such tasks would be a straightforward extension, and provide the possibility to include many more tasks with shared structure.
Another challenge when deploying robot learning and meta-learning algorithms is the manual effort of resetting the environment. To simulate this case, one simple extension of the benchmark is to significantly reduce the frequency of resets available to the robot while learning. Lastly, in many real-world situations, the tasks are not available all at once. To reflect this challenge in the benchmark, we can add an evaluation protocol that matches that of online meta-learning problem
statements~\cite{finn2019online}.
We leave these directions for future work, either to be done by ourselves or in the form of open-source contributions. 
To summarize, we believe that the proposed form of the task suite represents a significant step towards evaluating multi-task and meta-learning algorithms on diverse robotic manipulation problems that will pave the way for future research in these areas.



\acknowledgments{We thank Suraj Nair for feedback on a draft of the paper. We thank K.R Zentner for her help in maintaining Meta-World. This research was supported in part by the National Science Foundation under IIS-1651843, IIS-1700697, and IIS-1700696, the Office of Naval Research, ARL DCIST CRA W911NF-17-2-0181, DARPA, Google, Amazon, and NVIDIA.
}



\bibliography{references}  

\clearpage

\appendix
\begin{sidewaysfigure}
    \includegraphics[width=\columnwidth]{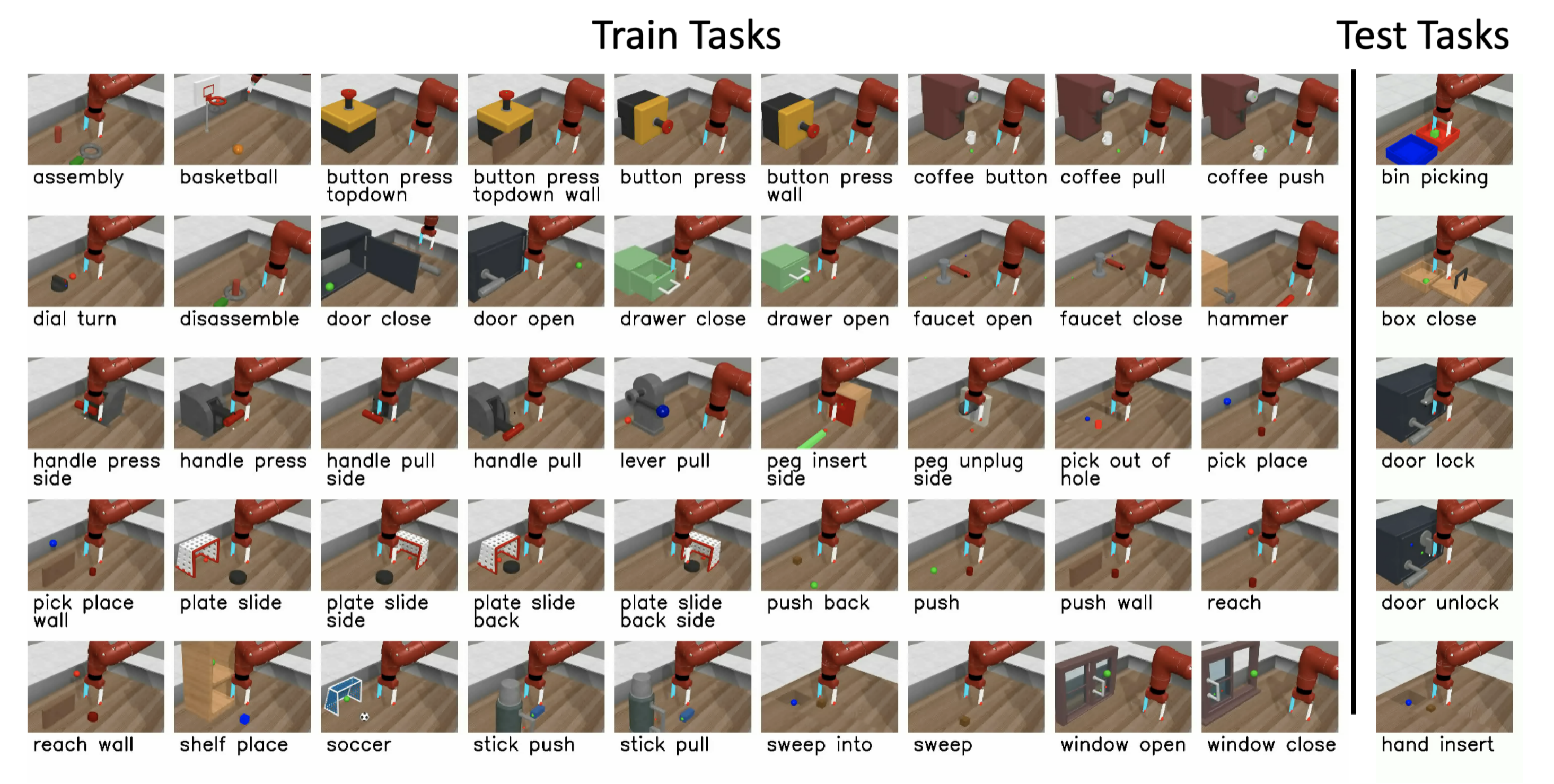}
    \caption{Enlarged image of Figure \ref{fig:ml45_teaser}.}
    \label{app:ML45_figure}
\end{sidewaysfigure}
\clearpage
\begin{sidewaysfigure}[p]
    \includegraphics[width=\columnwidth]{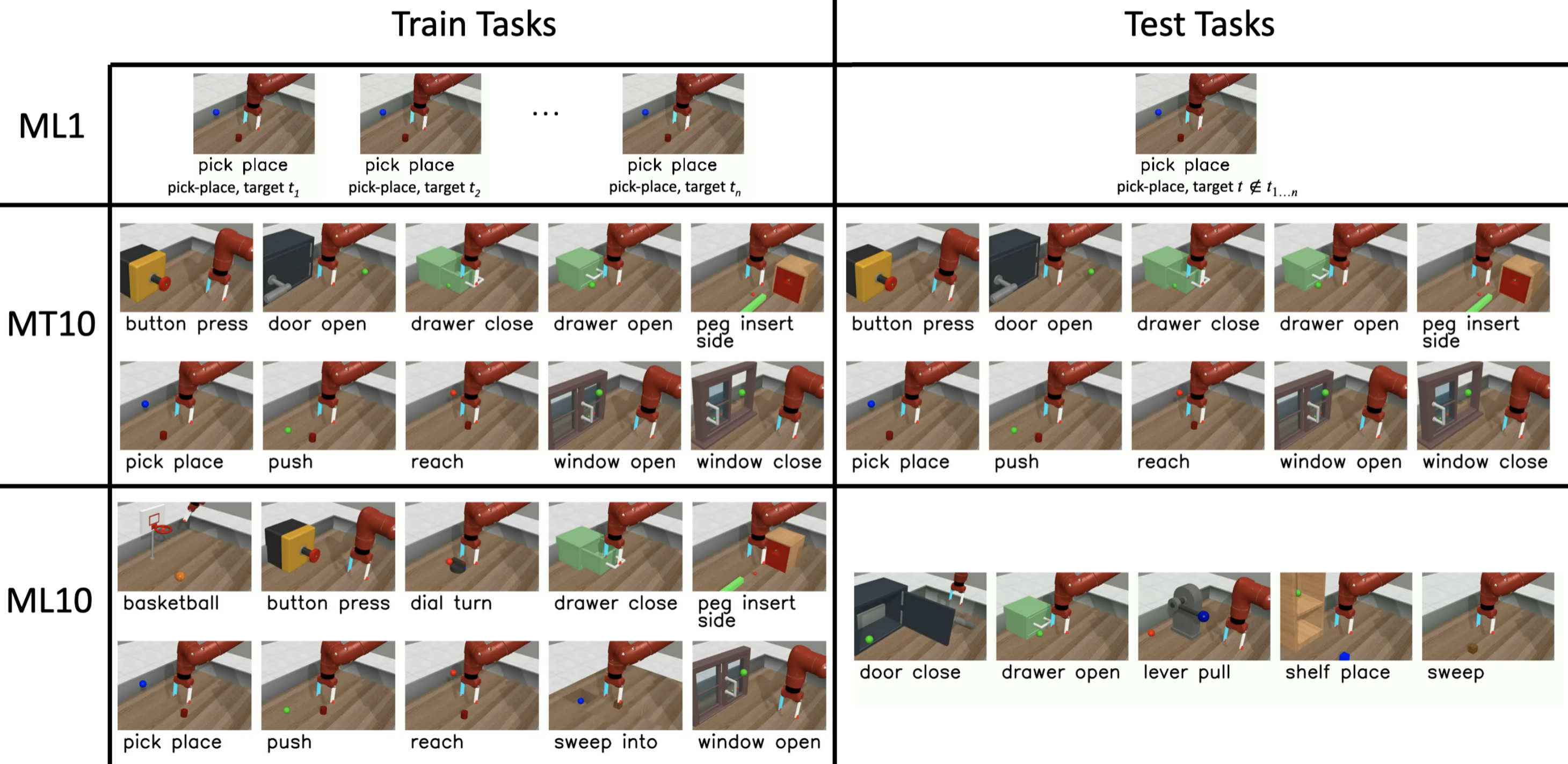}
    \caption{\footnotesize Enlarged image of Figure~\ref{fig:evaluation}.
    }
\end{sidewaysfigure}
\clearpage

\section{Task Descriptions}
\label{app:tasks}

In Table~\ref{tbl:tasks}, we include a description of each of the 50 Meta-World tasks.

\begin{table}[h]
\footnotesize
    \centering
    \begin{tabular}{ll}
\toprule
 Task & Description  \\
\midrule
turn on faucet & Rotate the faucet counter-clockwise. Randomize faucet positions \\
sweep & Sweep a puck off the table. Randomize puck positions\\
assemble nut & Pick up a nut and place it onto a peg. Randomize nut and peg positions\\
turn off faucet & Rotate the faucet clockwise. Randomize faucet positions\\
push & Push the puck to a goal. Randomize puck and goal positions\\
pull lever & Pull a lever down $90$ degrees. Randomize lever positions\\
turn dial & Rotate a dial $180$ degrees. Randomize dial positions\\
push with stick & Grasp a stick and push a box using the stick. Randomize stick positions.\\
get coffee & Push a button on the coffee machine. Randomize the position of the coffee machine\\
pull handle side & Pull a handle up sideways. Randomize the handle positions\\
basketball & Dunk the basketball into the basket. Randomize basketball and basket positions\\
pull with stick & Grasp a stick and pull a box with the stick. Randomize stick positions\\
sweep into hole & Sweep a puck into a hole. Randomize puck positions\\
disassemble nut & pick a nut out of the a peg. Randomize the nut positions\\
place onto shelf & pick and place a puck onto a shelf. Randomize puck and shelf positions\\
push mug & Push a mug under a coffee machine. Randomize the mug and the machine positions\\
press handle side & Press a handle down sideways. Randomize the handle positions\\
hammer & Hammer a screw on the wall. Randomize the hammer and the screw positions\\
slide plate & Slide a plate into a cabinet. Randomize the plate and cabinet positions\\
slide plate side & Slide a plate into a cabinet sideways. Randomize the plate and cabinet positions\\
press button wall & Bypass a wall and press a button. Randomize the button positions\\
press handle & Press a handle down. Randomize the handle positions\\
pull handle & Pull a handle up. Randomize the handle positions\\
soccer & Kick a soccer into the goal. Randomize the soccer and goal positions\\
retrieve plate side & Get a plate from the cabinet sideways. Randomize plate and cabinet positions\\
retrieve plate & Get a plate from the cabinet. Randomize plate and cabinet positions\\
close drawer & Push and close a drawer. Randomize the drawer positions\\
press button top & Press a button from the top. Randomize button positions\\
reach & reach a goal position. Randomize the goal positions\\
press button top wall & Bypass a wall and press a button from the top. Randomize button positions\\
reach with wall & Bypass a wall and reach a goal. Randomize goal positions\\
insert peg side & Insert a peg sideways. Randomize peg and goal positions\\
pull & Pull a puck to a goal. Randomize puck and goal positions\\
push with wall & Bypass a wall and push a puck to a goal. Randomize puck and goal positions\\
pick out of hole & Pick up a puck from a hole. Randomize puck and goal positions\\
pick\&place w/ wall & Pick a puck, bypass a wall and place the puck. Randomize puck and goal positions\\
press button & Press a button. Randomize button positions\\
pick\&place & Pick and place a puck to a goal. Randomize puck and goal positions\\
pull mug & Pull a mug from a coffee machine. Randomize the mug and the machine positions\\
unplug peg & Unplug a peg sideways. Randomize peg positions\\
close window & Push and close a window. Randomize window positions\\
open window & Push and open a window. Randomize window positions\\
open door & Open a door with a revolving joint. Randomize door positions\\
close door & Close a door with a revolving joint. Randomize door positions\\
open drawer & Open a drawer. Randomize drawer positions\\
insert hand & Insert the gripper into a hole.\\
close box & Grasp the cover and close the box with it. Randomize the cover and box positions\\
lock door & Lock the door by rotating the lock clockwise. Randomize door positions\\
unlock door & Unlock the door by rotating the lock counter-clockwise. Randomize door positions\\
pick bin & Grasp the puck from one bin and place it into another bin. Randomize puck positions\\
\bottomrule
\end{tabular}
\vspace{0.2cm}
    \caption{A list of all of the Meta-World tasks and a description of each task.}
    \label{tbl:tasks}
\end{table}

\section{Benchmark Verification with Single-Task Learning}
\label{app:singletask}

\begin{figure}[t]
    \centering
    \includegraphics[width=\columnwidth, left]{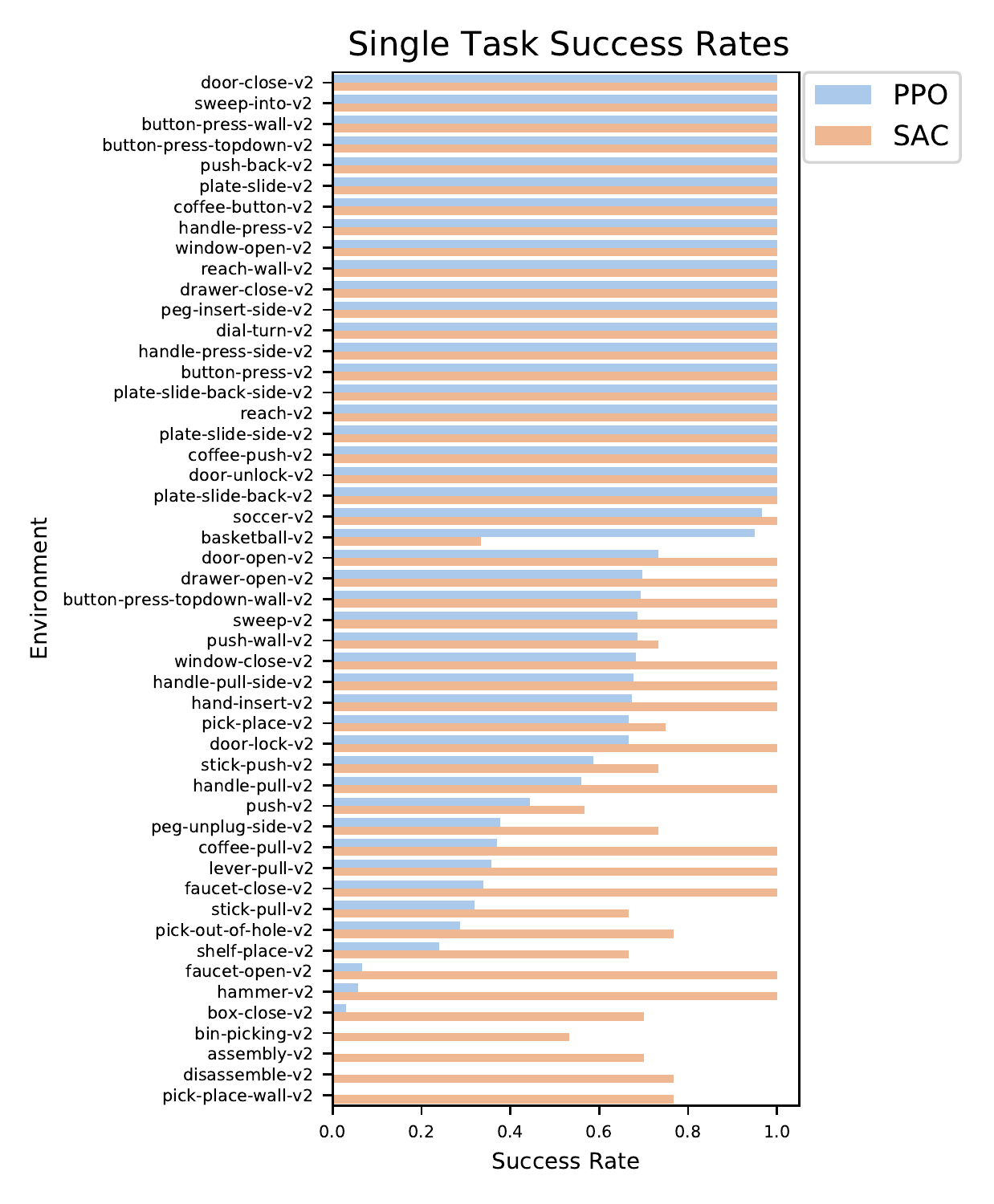}
    \vspace{-1cm}
    \caption{Performance of independent policies trained on individual tasks using soft actor-critic (SAC) and proximal policy optimization (PPO) on 3 seeds. We verify that SAC can solve all of the tasks and PPO can also solve most of the tasks.}
    \vspace{-0.5cm}
    \label{fig:single_task_results}
\end{figure}

In this section, we aim to verify that each of the benchmark tasks are individually solvable provided enough data. To do so, we consider two state-of-the-art single task reinforcement learning methods, proximal policy optimization (PPO)~\cite{schulman2017proximal} and soft actor-critic (SAC)~\cite{haarnoja2018soft}.
This evaluation is purely for validation of the tasks, and not an official evaluation protocol of the benchmark. Details of the hyperparameters are provided in Appendix~\ref{app:hyperparameters}.
The results of this experiment are illustrated in Figure~\ref{fig:single_task_results}. We indeed find that SAC can learn to perform all of the $50$ tasks to some degree, while PPO can solve a large majority of the tasks.

\section{Learning curves}
\label{app:curves}

\begin{figure}[H]
    \centering
    \includegraphics[width=0.9\columnwidth]{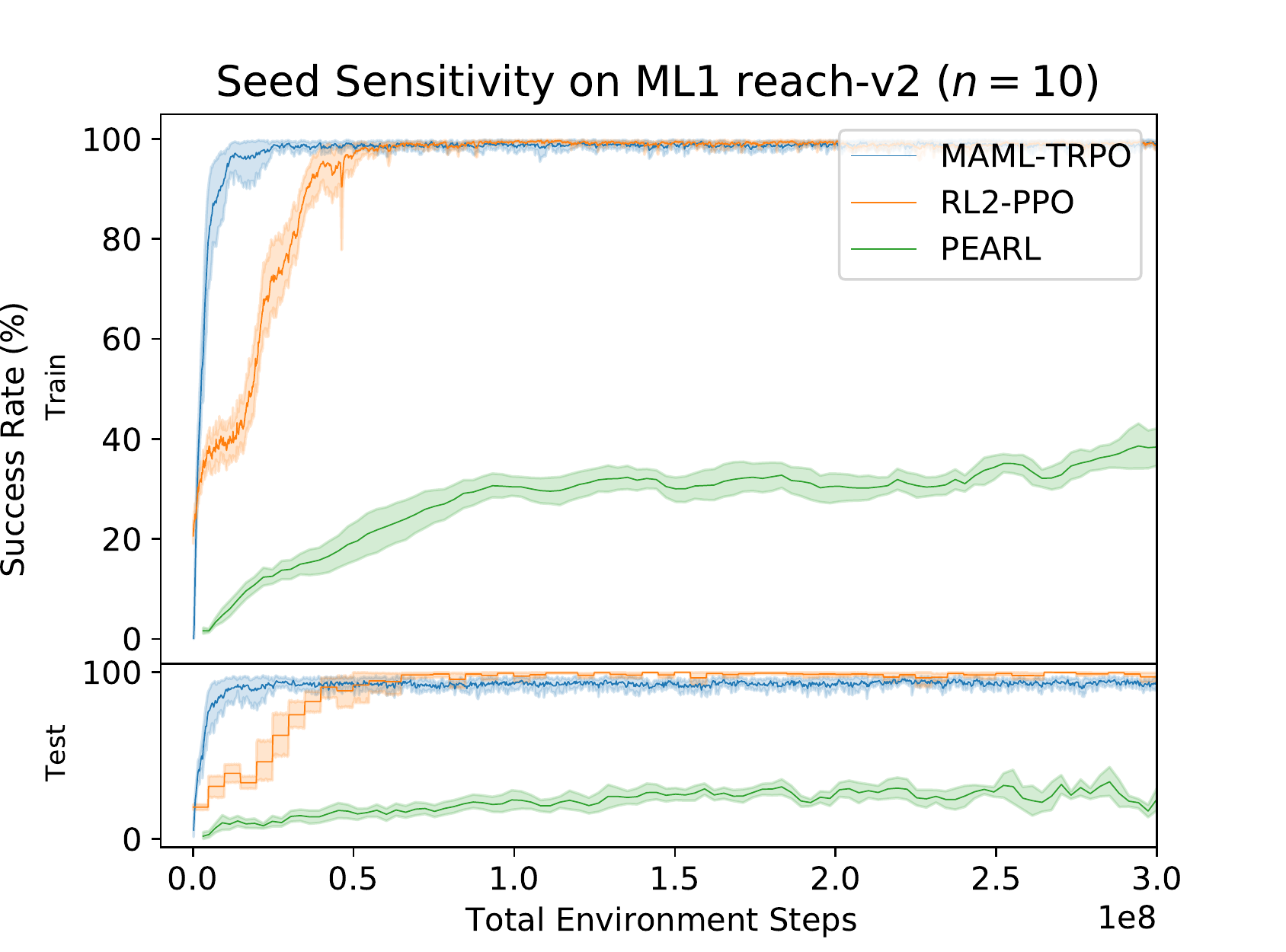}
    \caption{Comparison of PEARL, MAML, and RL$^2$ learning curves on ML-1 reach.}
    \label{fig:babyresults}
\end{figure}

\begin{figure}[H]
    \centering
    \includegraphics[width=0.9\columnwidth]{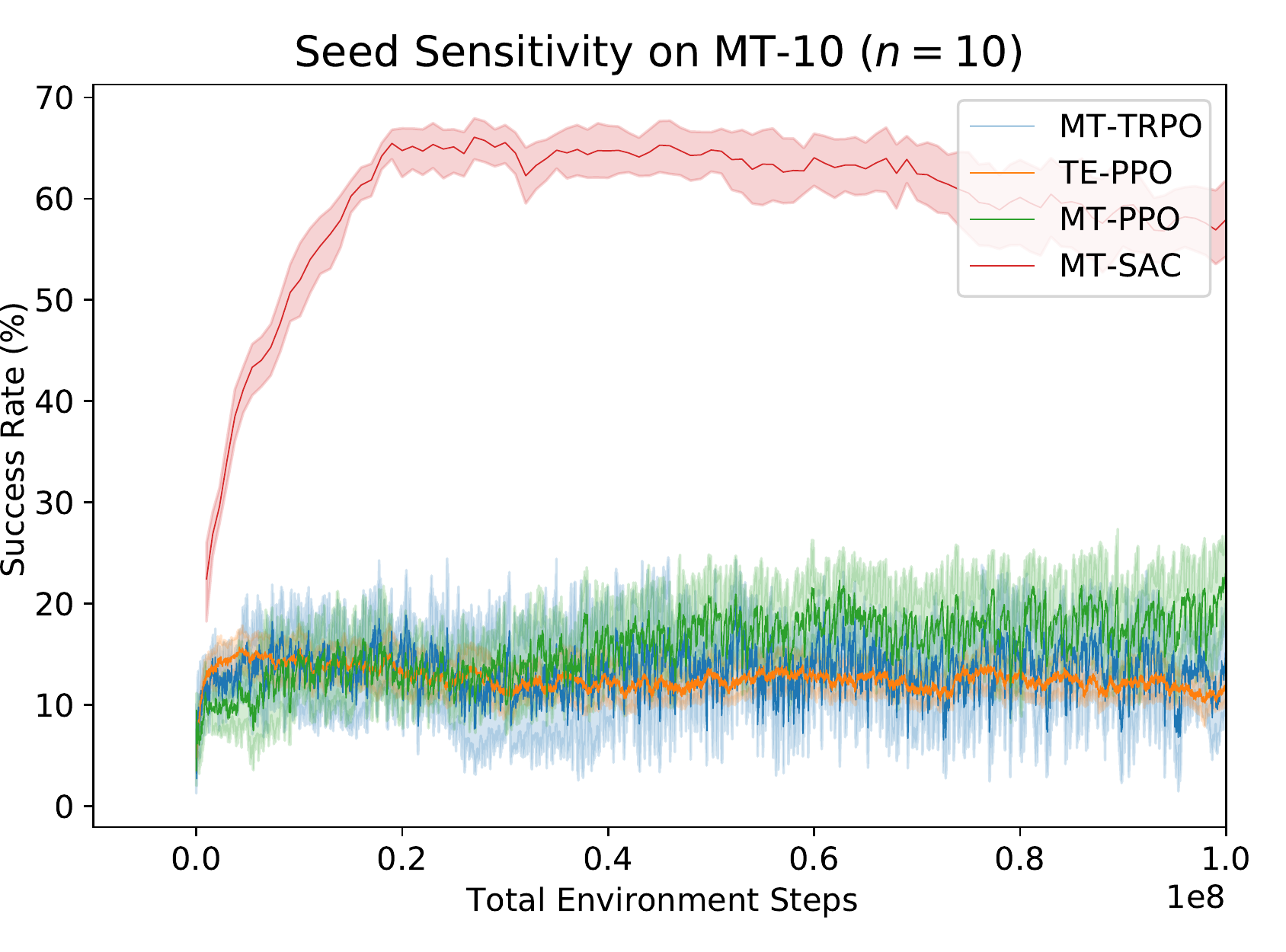}
    \caption{Comparison of MTRL algorithms on MT-10. MT-SAC vastly outperforms is on-policy counterparts in performance and sample efficiency.}
    \label{fig:learningcurves}
\end{figure}

In evaluating meta-learning algorithms, we care not just about performance but also about efficiency, i.e. the amount of data required by the meta-training process. While the adaptation process for all algorithms is extremely efficient, requiring only a few trajectories, the meta-learning process can be very inefficient.
In Figure~\ref{fig:babyresults}, we show full learning curves of the three meta-learning methods on ML1. In Figure~\ref{fig:learningcurves}, we show full learning curves of MT10, ML10, MT50 and ML45.
The MT10 and MT50 learning curves show the efficiency of multi-task learning, a critical evaluation metric, since sample efficiency gains are a primary motivation for using multi-task learning. 
Unsurprisingly, we find that off-policy algorithms such as soft actor-critic are able to learn with substantially less data than on-policy algorithms.

\begin{figure}[H]
    \centering
    \includegraphics[width=0.9\columnwidth]{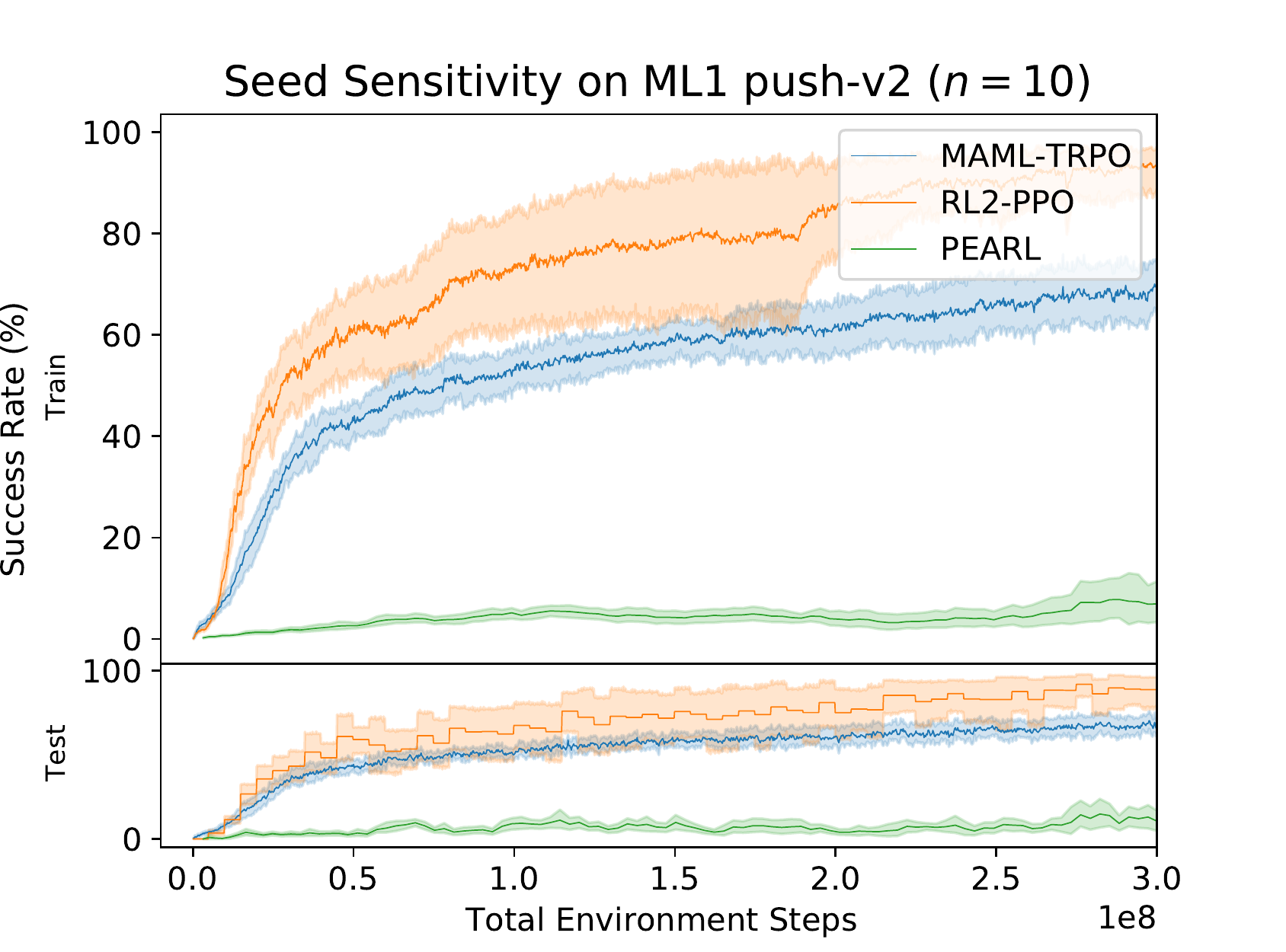}
    \caption{Comparison of PEARL, MAML, and RL$^2$ learning curves on ML-1 push.}
    \label{fig:ml1-push}
\end{figure}
\clearpage
\begin{figure}[H]
    \centering
    \includegraphics[width=0.9\columnwidth]{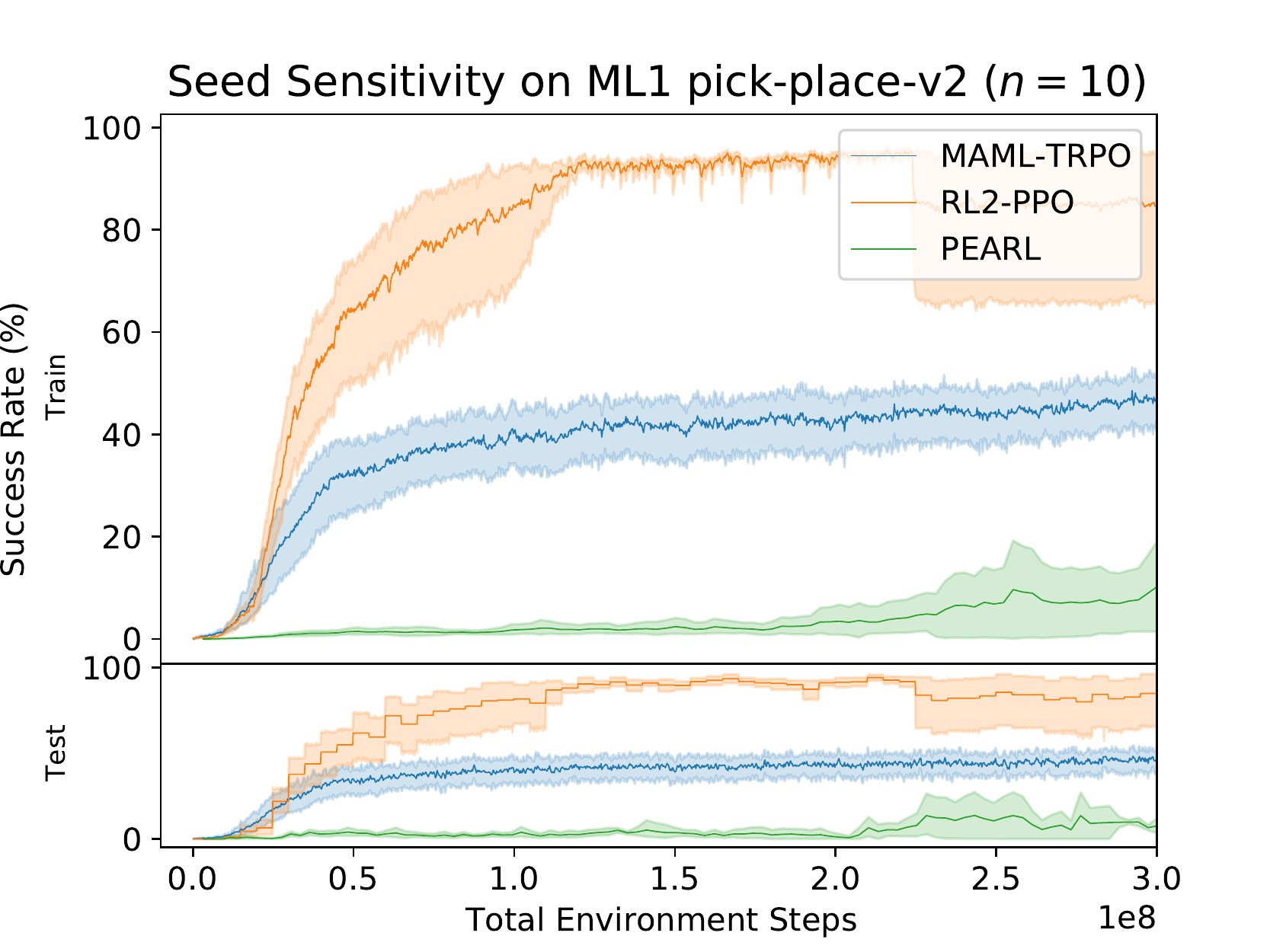}
    \label{fig:ml1-pick-place}
    \caption{Comparison of PEARL, MAML, and RL$^2$ learning curves on the simplest evaluation, ML-1, where the methods need to adapt quickly to new object and goal positions within the one meta-training task.}
\end{figure}

\begin{figure}[H]
    \centering
    \includegraphics[width=0.9\columnwidth]{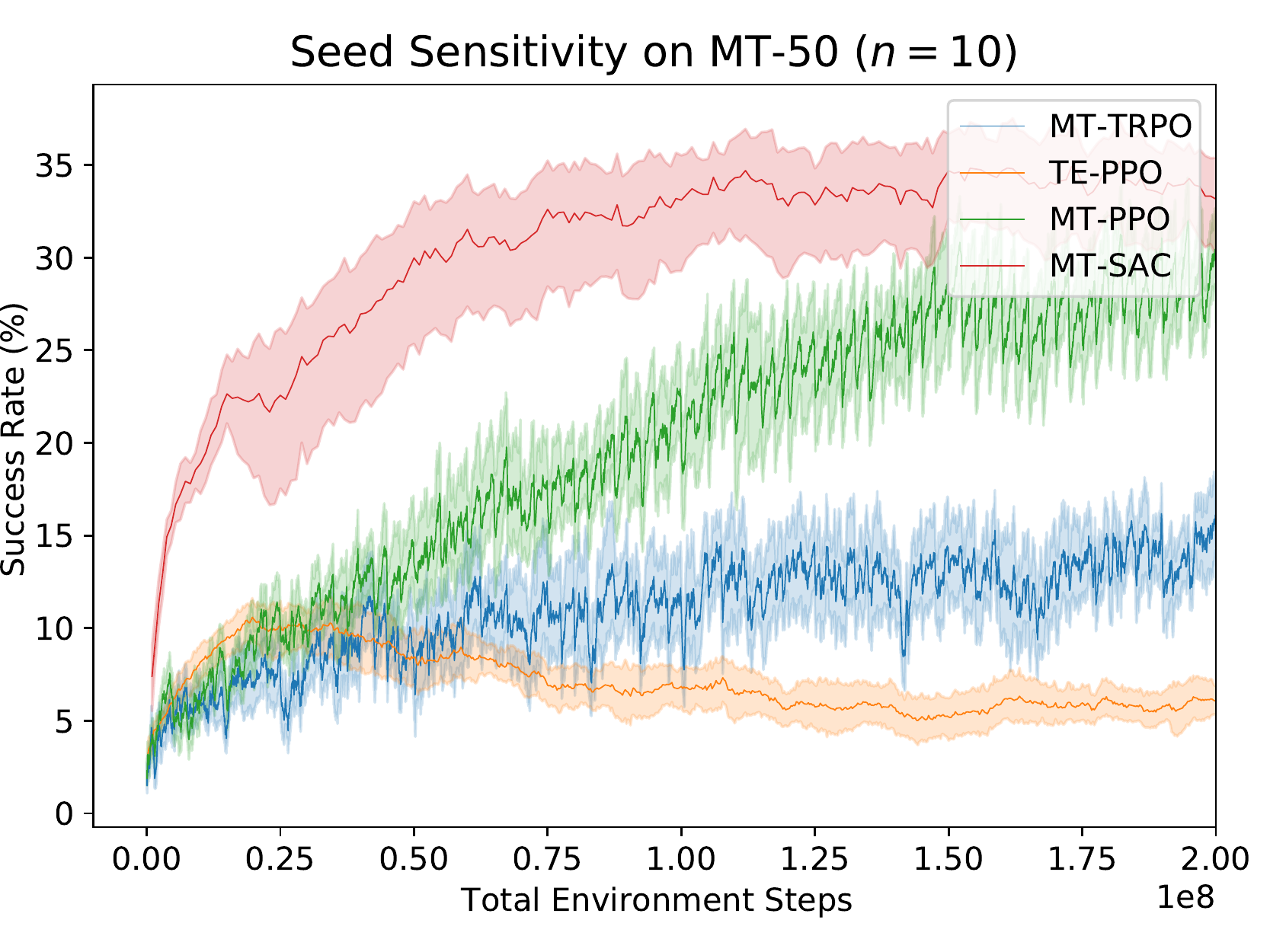}
    \vspace{-0.5cm}
    \caption{Comparison of MTRL algorithms on MT-50. MT-SAC vastly outperforms is on-policy counterparts in sample efficiency. Its performance tapers off, and with more training, MT-PPO outperforms it.}
    \label{fig:mt50-curve}
\end{figure}
\begin{figure}[H]
    \vspace{-2.5cm}
    \centering
    \includegraphics[width=\columnwidth]{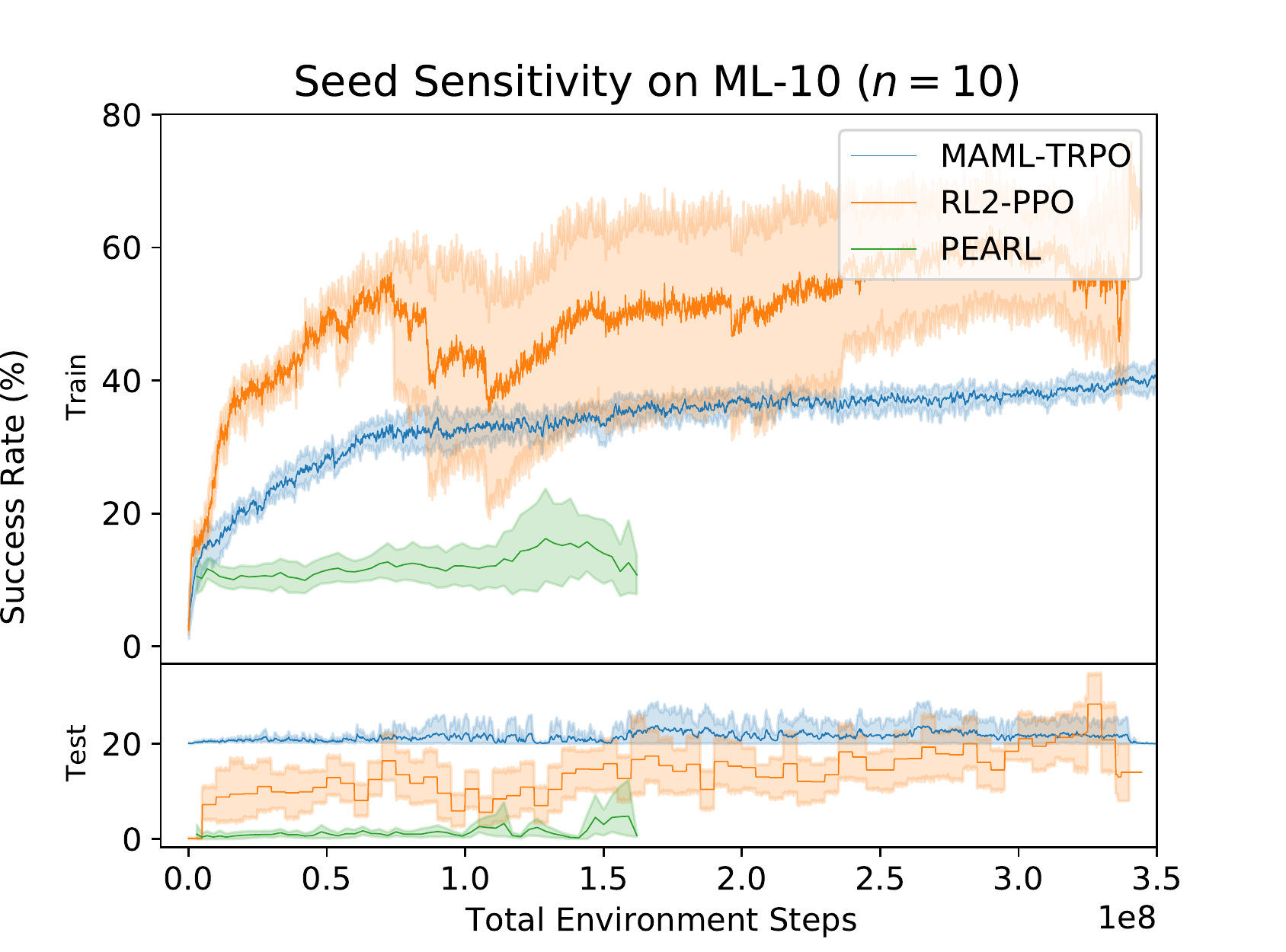}
    \caption{Performance of meta-RL algorithms on ML-10. RL$^2$ significantly outperforms other methods in terms of sample efficiency and performance on test tasks. MAML has better test performance early on, RL$^2$ outperforms it with more training.}
    \label{fig:ml10-curve}
\end{figure}
\begin{figure}[H]
    \centering
    \includegraphics[width=\columnwidth]{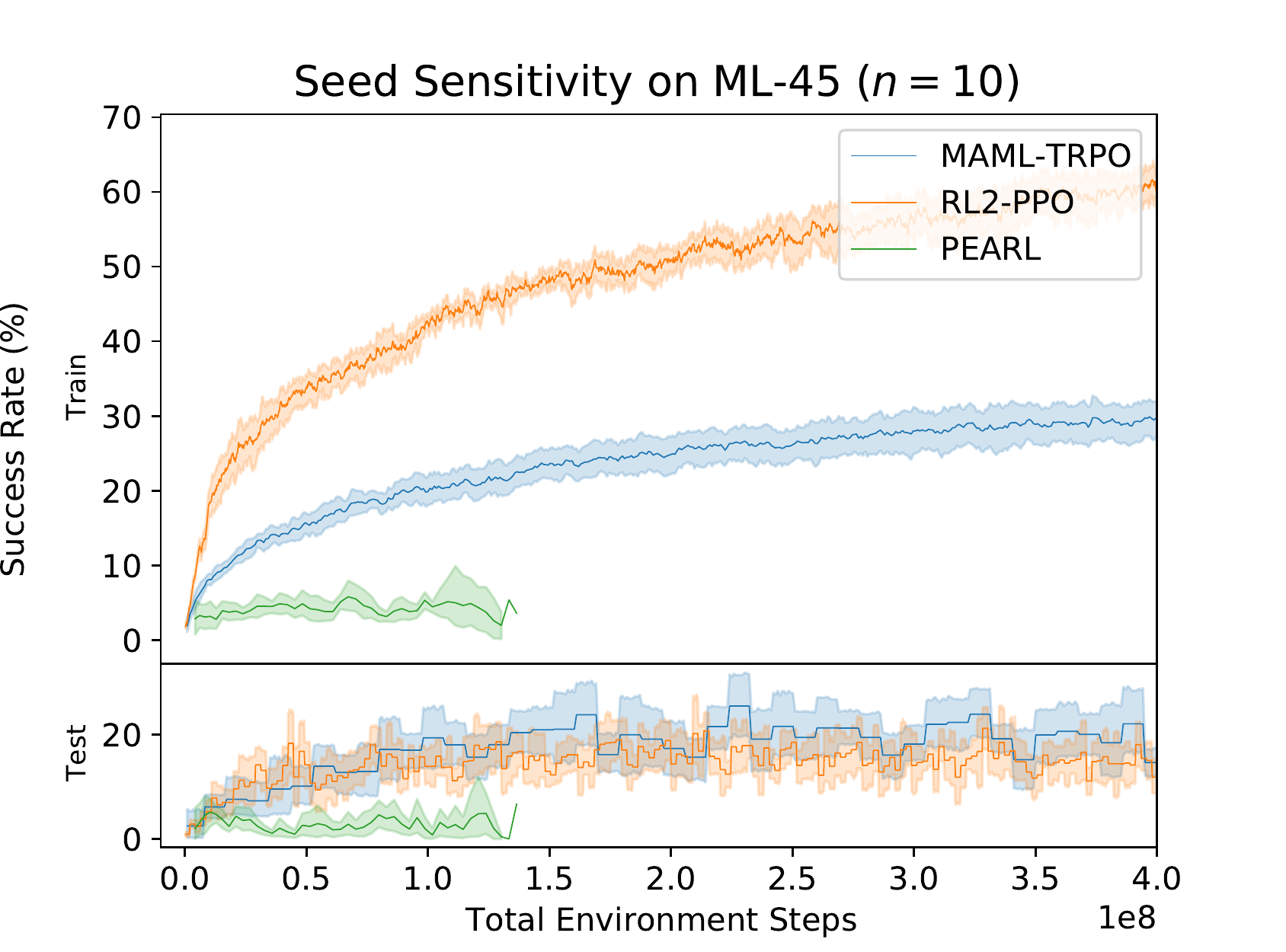}
    \caption{Learning curves of all methods on the ML-45 benchmark. Y-axis represents success rate averaged over tasks in percentage (\%). The dashed lines represent asymptotic performances. PEARL underperforms MAML and RL$^2$. RL$^2$ significantly outperforms other methods in terms of sample efficiency and performance on train tasks. RL$^2$ and MAML have similar performance on test tasks.}
    \label{fig:ml45-curve}
\end{figure}
\pagebreak{}

\section{Hyperparameter Details}
\label{app:hyperparameters}

In this section, we provide hyperparameter values for each of the methods in our experimental evaluation.

\subsection{Single Task SAC}

\begin{table}[h!]
\begin{tabularx}{\linewidth}{ L{1.25in} X X X l }
    \toprule
    \textbf{Description} & \textbf{value} & \texttt{variable\_name} \\
    \midrule
    \multicolumn{3}{l}{Normal Hyperparameters} \\
    \midrule
    Batch size & $500$ & \texttt{batch\_size} \\
    Number of epochs & $500$ & \texttt{n\_epochs} \\
    Path length per roll-out & $500$ & \texttt{max\_path\_length} \\
    Discount factor & $0.99$ & \texttt{discount} \\
    \midrule
    \multicolumn{3}{l}{Algorithm-Specific Hyperparameters} \\
    \midrule
    Policy hidden sizes & {\scriptsize $(256, 256)$}& \texttt{hidden\_sizes} \\
    Activation function of hidden layers & ReLU & \texttt{hidden\_nonlinearity} \\
    Policy learning rate & \num{3e-4} & \texttt{policy\_lr} \\
    Q-function learning rate & \num{3e-4} & \texttt{qf\_lr} \\
    Policy minimum standard deviation & \(e^{-20}\) & \texttt{min\_std} \\
    Policy maximum standard deviation & \(e^{2}\) & \texttt{max\_std} \\
    Gradient steps per epoch & 500 & \texttt{gradient\_steps\_per\_itr} \\
    Number of epoch cycles & 40 & \texttt{epoch\_cycles} \\
    Soft target interpolation parameter & \num{5e-3} & \texttt{target\_update\_tau} \\
    Use automatic entropy Tuning & True & \texttt{use\_automatic\_entropy\_tuning} \\
    \bottomrule
\end{tabularx}
\caption{Hyperparameters used for Garage experiments with Single Task SAC}
\label{tab:garage_sac_hparams}
\end{table}
\clearpage
\subsection{Single Task PPO}

\begin{table}[h!]
\begin{tabularx}{\linewidth}{ L{1.5in} X X X l }
    \toprule
    \textbf{Description} & \textbf{value} & \texttt{variable\_name} \\
    \midrule
    \multicolumn{3}{l}{Normal Hyperparameters} \\
    \midrule
    Batch size & $5{,}000$ & \texttt{batch\_size} \\
    Number of epochs & $4{,}000$ & \texttt{n\_epochs} \\
    Path length per roll-out & $500$ & \texttt{max\_path\_length} \\
    Discount factor & $0.99$ & \texttt{discount} \\
    \midrule
    \multicolumn{3}{l}{Algorithm-Specific Hyperparameters} \\
    \midrule
    Policy mean hidden sizes & $(128, 128)$ & \texttt{hidden\_sizes} \\
    Policy minimum standard deviation & $0.5$ & \texttt{min\_std} \\
    Policy maximum standard deviation & $1.5$ & \texttt{max\_std} \\
    Policy share standard deviation and mean network & True & \texttt{std\_share\_network} \\
    Activation function of mean hidden layers & tanh & \texttt{hidden\_nonlinearity} \\
    Optimizer learning rate & \num{5e-4} & \texttt{learning\_rate} \\ 
    Likelihood ratio clip range & $0.2$ & \texttt{lr\_clip\_range} \\
    Advantage estimation $\lambda$ & $0.95$ & \texttt{gae\_lambda} \\
    Use layer normalization & False & \texttt{layer\_normalization} \\
    Entropy method &  \detokenize{max} & \texttt{entropy\_method} \\
    Loss function &  \detokenize{surrogate_clip} & \texttt{pg\_loss} \\
    Maximum number of epochs for update & $256$ & \texttt{max\_epochs} \\
    Minibatch size for optimization & $32$ & \texttt{batch\_size} \\
    \midrule
    \multicolumn{3}{l}{Value Function Hyperparameters} \\
    \midrule
    Policy hidden sizes & $(128, 128)$ & \texttt{hidden\_sizes} \\
    Activation function of hidden layers & tanh & \texttt{hidden\_nonlinearity} \\
    Initial value for standard deviation & $1$ & \texttt{init\_std} \\
    Use trust region constraint & False & \texttt{use\_trust\_region} \\
    Normalize inputs & True & \texttt{normalize\_inputs} \\
    Normalize outputs & True & \texttt{normalize\_outputs} \\
    \bottomrule
\end{tabularx}
\caption{Hyperparameters used for Garage experiments with Single Task PPO}
\label{tab:garage_st_ppo_hparams}
\end{table}

Below we summarize in as much detail as possible the hyperparameters used for each experiment in this chapter.
Seed values were individually chosen at random for each experiment.

\clearpage
\subsection{MT-PPO}

\begin{table}[h!]
\begin{tabularx}{\linewidth}{ L{1.7in} X X X l }
    \toprule
    \textbf{Description} & \textbf{MT10} & \textbf{MT50} & \texttt{variable\_name} \\
    \midrule
    \multicolumn{4}{l}{Normal Hyperparameters} \\
    \midrule
    Batch size & $100{,}000$ & $500{,}000$ & \texttt{batch\_size} \\
    Number of epochs & $10{,}000$ & $10{,}000$ & \texttt{n\_epochs} \\
    Path length per roll-out & $500$ & $500$ & \texttt{max\_path\_length} \\
    Discount factor & $0.99$ & $0.99$ & \texttt{discount} \\
    \midrule
    \multicolumn{4}{l}{Algorithm-Specific Hyperparameters} \\
    \midrule
    Policy mean hidden sizes & $(512, 512)$ & \texttt{hidden\_sizes} \\
    Policy minimum standard deviation & $0.5$ & $0.5$ & \texttt{min\_std} \\
    Policy maximum standard deviation & $1.5$ & $1.5$ & \texttt{max\_std} \\
    Policy share standard deviation and mean network & True & True & \texttt{std\_share\_network} \\
    Activation function of hidden layers & tanh & tanh & \texttt{hidden\_nonlinearity} \\
    Optimizer learning rate & \num{5e-4} & \num{5e-4} & \texttt{learning\_rate} \\ 
    Likelihood ratio clip range & $0.2$ & $0.2$ & \texttt{lr\_clip\_range} \\
    Advantage estimation $\lambda$ & $0.97$ & $0.97$ & \texttt{gae\_lambda} \\
    Use layer normalization & False & False & \texttt{layer\_normalization} \\
    Use trust region constraint & False & False & \texttt{use\_trust\_region} \\
    Entropy method &  \detokenize{max} & \detokenize{max} & \texttt{entropy\_method} \\
    Policy entropy coefficient & $5e-3$ & $5e-3$ & \texttt{policy\_ent\_coeff} \\
    Loss function &  \detokenize{surrogate_clip} & \detokenize{surrogate_clip} & \texttt{pg\_loss} \\
    Maximum number of epochs for update & $16$ & $16$ & \texttt{max\_epochs} \\
    Minibatch size for optimization & $32$ & $32$ & \texttt{batch\_size} \\
    \midrule
    \multicolumn{4}{l}{Value Function Hyperparameters} \\
    \midrule
    Value Function hidden sizes & $(512, 512)$ & $(512, 512)$ & \texttt{hidden\_sizes} \\
    Activation function of hidden layers & tanh & tanh & \texttt{hidden\_nonlinearity} \\
    Trainable standard deviation  & True & True & \texttt{learn\_std} \\
    Initial value for standard deviation & $1$ & $1$ & \texttt{init\_std} \\
    Use layer normalization & False & False & \texttt{layer\_normalization} \\
    Use trust region constraint & False & False & \texttt{use\_trust\_region} \\
    Normalize inputs & True & True & \texttt{normalize\_inputs} \\
    Normalize outputs & True & True & \texttt{normalize\_outputs} \\
    \bottomrule
\end{tabularx}
\caption{Hyperparameters used for Garage experiments with Multi-Task PPO}
\label{tab:garage_ppo_hparams}
\end{table}

\FloatBarrier

\clearpage
\subsection{MT-TRPO}

\FloatBarrier

\begin{table}[h!]
\begin{tabularx}{\linewidth}{ L{2in} X X X l }
    \toprule
    \textbf{Description} & \textbf{MT10} & \textbf{MT50} & \texttt{variable\_name} \\
    \midrule
    \multicolumn{4}{l}{Normal Hyperparameters} \\
    \midrule
    Batch size & $100{,}000$ & $500{,}000$ & \texttt{batch\_size} \\
    Number of epochs & $10{,}000$ & $10{,}000$ & \texttt{n\_epochs} \\
    Path length per roll-out & $500$ & $500$ & \texttt{max\_path\_length} \\
    Discount factor & $0.99$ & $0.99$ & \texttt{discount} \\
    \midrule
    \multicolumn{4}{l}{Algorithm-Specific Hyperparameters} \\
    \midrule
    Policy mean hidden sizes & $(512, 512)$ & \texttt{hidden\_sizes} \\
    Policy minimum standard deviation & $0.5$ & $0.5$ & \texttt{min\_std} \\
    Policy maximum standard deviation & $1.5$ & $1.5$ & \texttt{max\_std} \\
    Policy share standard deviation and mean network & True & True & \texttt{std\_share\_network} \\
    Activation function of hidden layers & tanh & tanh & \texttt{hidden\_nonlinearity} \\
    Advantage estimation $\lambda$ & $0.95$ & $0.95$ & \texttt{gae\_lambda} \\
    Maximum KL divergence & \num{1e-2} & \num{1e-2} & \texttt{max\_kl\_step} \\
    Number of CG iterations & $10$ & $10$ & \texttt{cg\_iters} \\
    Regularization coefficient & \num{1e-5} & \num{1e-5} & \texttt{reg\_coeff} \\
    Use layer normalization & False & False & \texttt{layer\_normalization} \\
    Use trust region constraint & False & False & \texttt{use\_trust\_region} \\
    Entropy method & \detokenize{no_entropy} & \detokenize{no_entropy} & \texttt{entropy\_method} \\
    Loss function & surrogate & surrogate & \texttt{pg\_loss} \\
    \midrule
    \multicolumn{4}{l}{Value Function Hyperparameters} \\
    \midrule
    Hidden sizes & $(512, 512)$ & $(512, 512)$ & \texttt{hidden\_sizes} \\
    Activation function of hidden layers & tanh & tanh & \texttt{hidden\_nonlinearity} \\
    Trainable standard deviation & True & True & \texttt{learn\_std} \\
    Initial value for standard deviation & $1$ & $1$ & \texttt{init\_std} \\
    Use layer normalization & False & False & \texttt{layer\_normalization} \\
    Use trust region constraint & True & True & \texttt{use\_trust\_region} \\
    Normalize inputs & True & True & \texttt{normalize\_inputs} \\
    Normalize outputs & True & True & \texttt{normalize\_outputs} \\
    \bottomrule
\end{tabularx}
\caption{Hyperparameters used for Garage experiments with Multi-Task TRPO}
\label{tab:garage_trpo_hparams}
\end{table}

\FloatBarrier

\clearpage
\subsection{MT-SAC}

\FloatBarrier

\begin{table}[h!]
\begin{tabularx}{\linewidth}{ L{1.5in} X X X l }
    \toprule
    \textbf{Description} & \textbf{MT10} & \textbf{MT50} & \texttt{variable\_name} \\
    \midrule
    \multicolumn{4}{l}{General Hyperparameters} \\
    \midrule
    Batch size & $5{,}000$ & $25{,}000$ & \texttt{batch\_size} \\
    Number of epochs & $500$ & $500$ & \texttt{epochs} \\
    Path length per roll-out & $500$ & $500$ & \texttt{max\_path\_length} \\
    Discount factor & $0.99$ & $0.99$ & \texttt{discount} \\
    \midrule
    \multicolumn{4}{l}{Algorithm-Specific Hyperparameters} \\
    \midrule
    Policy hidden sizes & {\scriptsize $(400, 400)$} & {\scriptsize $(400, 400)$} & \texttt{hidden\_sizes} \\
    Activation function of hidden layers & ReLU & ReLU & \texttt{hidden\_nonlinearity} \\
    Policy learning rate & \num{3e-4} & \num{3e-4} & \texttt{policy\_lr} \\
    Q-function learning rate & \num{3e-4} & \num{3e-4} & \texttt{qf\_lr} \\
    Policy minimum standard deviation & \(e^{-20}\) & \(e^{-20}\) & \texttt{min\_std} \\
    Policy maximum standard deviation & \(e^{2}\) & \(e^{2}\) & \texttt{max\_std} \\
    Gradient steps per epoch & 500 & 500 & \texttt{gradient\_steps\_per\_itr} \\
    Number of epoch cycles & 200 & 40 & \texttt{epoch\_cycles} \\
    Soft target interpolation parameter & \num{5e-3} & \num{5e-3} & \texttt{target\_update\_tau} \\
    Use automatic entropy Tuning & True & True & \texttt{use\_automatic\_entropy\_tuning} \\
    Minimum Buffer Batch Size & 1500 & 7500 & \texttt{min\_buffer\_size} \\
    \bottomrule
\end{tabularx}
\caption{Hyperparameters used for Garage experiments with Multi-Task SAC}
\label{tab:garage_mtsac_hparams}
\end{table}

\FloatBarrier

\clearpage
\subsection{TE-PPO}

\FloatBarrier

\begin{table}[h!]
\begin{tabularx}{\linewidth}{ l X X X l }
    \toprule
    \textbf{Description} & \textbf{MT10} & \textbf{MT50} & \texttt{argument\_name} \\
    \midrule
    General Hyperparameters \\
    \midrule
    Batch size & $50{,}000$ & $250{,}000$ & \texttt{batch\_size} \\
    Number of epochs & $4{,}000$ & $2{,}000$ & \texttt{n\_epochs} \\
    \midrule
    Algorithm-Specific Hyperparameters \\
    \midrule
    Policy hidden sizes & $(32, 16)$ & $(32, 16)$ & \texttt{hidden\_sizes} \\
    Activation function of hidden layers & tanh & tanh & \texttt{hidden\_nonlinearity} \\
    Likelihood ratio clip range & $0.2$ & $0.2$ & \texttt{lr\_clip\_range} \\
    Latent dimension & $4$ & $4$ & \texttt{latent\_length} \\
    Inference window length & $6$ & $6$ & \texttt{inference\_window} \\ 
    Embedding maximum standard deviation & $0.2$ & $0.2$ & \texttt{embedding\_max\_std} \\
    Policy entropy coefficient & $2e-2$ & $2e-2$ & \texttt{policy\_ent\_coeff} \\
    Value function & \multicolumn{2}{L{4.2cm}}{Gaussian MLP fit with observations, latent variables and returns} & \texttt{baseline} \\
    \bottomrule
\end{tabularx}
\caption{Hyperparameters used for Garage experiments with Task Embeddings PPO}
\label{tab:garage_te_hparams}
\end{table}

\FloatBarrier

\clearpage

\subsection{MAML}

\FloatBarrier

\begin{table}[h!]
\begin{tabularx}{\linewidth}{ L{1.5in} X X X l }
    \toprule
    \textbf{Description} & \textbf{ML1} & \textbf{ML10} & \textbf{ML45} & \texttt{argument\_name} \\
    \midrule
    \multicolumn{5}{l}{Meta-/Multi-Task Hyperparameters} \\
    \midrule
    Meta-batch size & $20$ & $20$ & $45$ & \texttt{meta\_batch\_size} \\
    Roll-outs per task & $10$ & $10$ & $20$ & \texttt{rollouts\_per\_task} \\
    \midrule
    General Hyperparameters \\
    \midrule
    Path length per roll-out & $500$ & $500$ & $500$ & \texttt{max\_path\_length} \\
    Discount factor & $0.99$ & $0.99$ & $0.99$ & \texttt{discount} \\
    \midrule
    Algorithm-specific Hyperparameters \\
    \midrule
    Policy hidden sizes & $(128, 128)$ & $(128, 128)$ & $(128, 128)$ & \texttt{hidden\_sizes} \\
    Activation function of hidden layers & tanh & tanh & tanh &  \texttt{hidden\_nonlinearity} \\
    Activation function of output layer & tanh & tanh & tanh &  \texttt{output\_nonlinearity} \\
    Inner algorithm learning rate & \num{1e-4} & \num{1e-4} & \num{1e-4} & \texttt{inner\_lr} \\
    Optimizer learning rate & \num{1e-3} & \num{1e-3} & \num{1e-3} & \texttt{outer\_lr} \\ 
    Maximum KL divergence & \num{1e-2} & \num{1e-2} & \num{1e-2} & \texttt{max\_kl\_step} \\
    Number of inner gradient updates & $1$ & $1$ & $1$ & \texttt{num\_grad\_update} \\
    Policy entropy coefficient & \num{5e-5} & \num{5e-5} & \num{5e-5} & \texttt{policy\_ent\_coeff} \\
    \bottomrule
\end{tabularx}
\caption{Hyperparameters used for Garage experiments with MAML}
\label{tab:garage_maml_hparams}
\end{table}

\FloatBarrier

\clearpage
\subsection{\RLsq}

\FloatBarrier

\begin{table}[h!]
\begin{tabularx}{\linewidth}{ L{1.5in} X X X l }
    \toprule
    \textbf{Description} & \textbf{ML1} & \textbf{ML10} & \textbf{ML45} & \texttt{argument\_name} \\
    \midrule
    \multicolumn{5}{l}{Meta-/Multi-Task Hyperparameters} \\
    \midrule
    Meta-batch size & $25$ & $10$ & $25$ & \texttt{meta\_batch\_size} \\
    Roll-outs per task & $10$ & $10$ & $10$ & \texttt{rollouts\_per\_task} \\    
    \midrule
    \multicolumn{5}{l}{General Hyperparameters} \\
    \midrule
    Path length per roll-out & $500$ & $500$ & $500$ & \texttt{max\_path\_length} \\
    Discount factor & $0.99$ & $0.99$ & $0.99$ & \texttt{discount} \\
    \midrule
    \multicolumn{5}{l}{Algorithm-Specific Hyperparameters} \\
    \midrule
    Policy hidden sizes & {\scriptsize $(256, )$} & {\scriptsize $(256, )$} & {\scriptsize $(256, )$} & \texttt{hidden\_sizes} \\
    Activation function of hidden layers & tanh & tanh & tanh & \texttt{hidden\_nonlinearity} \\
    Activation function of recurrent layers & sigmoid & sigmoid & sigmoid & \texttt{recurrent\_nonlinearity} \\
    Optimizer learning rate & \num{5e-4} & \num{5e-4} & \num{5e-4} & \texttt{optimizer\_lr} \\
    Likelihood ratio clip range & $0.2$ & $0.2$ & $0.2$ & \texttt{lr\_clip\_range} \\
    Advantage estimation $\lambda$ & $0.95$ & $0.95$ & $0.95$ & \texttt{gae\_lambda} \\
    Optimizer maximum epochs & $10$ & $10$ & $10$ & \texttt{optimizer\_max\_epochs} \\
    RNN cell type used in Policy & GRU & GRU & GRU & \texttt{cell\_type} \\
    Value function & \multicolumn{3}{C{5cm}}{Linear feature baseline} & \texttt{baseline} \\
    Policy entropy coefficient & \num{5e-6} & \num{5e-6} & \num{5e-6} & \texttt{policy\_ent\_coeff} \\
    Minimum policy standard deviation & $0.5$ & $0.5$ & $0.5$ & \texttt{min\_std} \\
    Maximum policy standard deviation & $0.5$ & $0.5$ & $0.5$ & \texttt{max\_std} \\ 
    \bottomrule
\end{tabularx}
\caption{Hyperparameters used for Garage experiments with \RLsq}
\label{tab:garage_rl2_hparams}
\end{table}

\FloatBarrier

\clearpage
\subsection{PEARL}

\FloatBarrier

\begin{table}[h!]
\begin{tabularx}{\linewidth}{ L{1.5in} X X X l }
    \toprule
    \textbf{Description} & \textbf{ML1} & \textbf{ML10} & \textbf{ML45} & \texttt{argument\_name} \\
    \midrule
    \multicolumn{5}{l}{Meta-/Multi-Task Hyperparameters} \\
    \midrule
    Meta-batch size & $16$ & $10$ &  $45$ & \texttt{meta\_batch\_size} \\
    Tasks sampled per epoch & $15$ & $10$ &  $45$ & \texttt{num\_tasks\_sample} \\
    Number of independent evaluations & \multicolumn{3}{C{3cm}}{$5$} & \texttt{num\_evals} \\
    Steps sampled per evaluation & $450$ & $1{,}650$ & $1{,}650$ & \texttt{num\_steps\_per\_eval} \\
    \midrule
    General Hyperparameters \\
    \midrule
    Batch size & $500$ & $1{,}000$ & $1{,}000$ &\texttt{batch\_size} \\
    Path length per roll-out & \multicolumn{3}{C{3cm}}{$500$} & \texttt{max\_path\_length} \\
    Reward scale & \multicolumn{3}{C{3cm}}{$10{,}000$} & \texttt{reward\_scale} \\
    Discount factor & \multicolumn{3}{C{3cm}}{$0.99$} & \texttt{discount} \\
    \midrule
    Algorithm-Specific Hyperparameters \\
    \midrule
    Policy hidden sizes & \multicolumn{3}{C{3cm}}{$(300, 300, 300)$} & \texttt{net\_size} \\
    Activation function of hidden layers & \multicolumn{3}{C{3cm}}{ReLU} & \texttt{hidden\_nonlinearity} \\
    Policy learning rate & \multicolumn{3}{C{3cm}}{ \num{3e-4}} & \texttt{policy\_lr} \\
    Q-function learning rate & \multicolumn{3}{C{3cm}}{ \num{3e-4}} & \texttt{qf\_lr} \\
    Value function learning rate & \multicolumn{3}{C{3cm}}{ \num{3e-4}} & \texttt{vf\_lr} \\
    Context learning rate & \multicolumn{3}{C{3cm}}{ \num{3e-4}} & \texttt{context\_lr} \\
    Latent dimension & \multicolumn{3}{C{3cm}}{$7$} & \texttt{latent\_dimension} \\
    Policy mean regularization coefficient & \multicolumn{3}{C{3cm}}{ \num{1e-3}} & \texttt{policy\_mean\_reg\_coeff} \\
    Policy standard deviation regularization coefficient & \multicolumn{3}{C{3cm}}{ \num{1e-3}} & \texttt{policy\_std\_reg\_coeff} \\
    Soft target interpolation parameter & \multicolumn{3}{C{3cm}}{ \num{5e-3}} & \texttt{soft\_target\_tau} \\
    KL $\lambda$ & \multicolumn{3}{C{3cm}}{$0.01$} & \texttt{KL\_lambda} \\
    Use information bottleneck & \multicolumn{3}{C{3cm}}{True} & \texttt{use\_information\_bottleneck} \\
    Use next observation in context & \multicolumn{3}{C{3cm}}{False} & \texttt{use\_next\_observation\_in\_context} \\
    Gradient steps per epoch & $1$ & $60$ & $14$ & \texttt{num\_steps\_per\_epoch} \\
    Steps sampled in the initial epoch & $1{,}000$ & $5{,}000$ & $22{,}500$ & \texttt{num\_initial\_steps} \\
    Prior steps sampled per epoch  & \multicolumn{3}{C{3cm}}{$2500$} & \texttt{num\_steps\_prior} \\
    Posterior steps sampled per epoch  & \multicolumn{3}{C{3cm}}{$2500$} & \texttt{num\_steps\_posterior} \\
    Extra posterior steps sampled per epoch  & \multicolumn{3}{C{3cm}}{$2500$} & \texttt{num\_extra\_steps\_posterior} \\
    Embedding batch size & \multicolumn{3}{C{3cm}}{$250$} & \texttt{embedding\_batch\_size} \\
    Embedding minibatch size & \multicolumn{3}{C{3cm}}{$250$} & \texttt{embedding\_mini\_batch\_size} \\
    \bottomrule
\end{tabularx}
\caption{Hyperparameters used for Garage experiments with PEARL}
\label{tab:garage_pearl_hparams}
\vspace{0.75in}
\end{table}

\FloatBarrier

\setttsize{}

\pagebreak

\section{Reward Functions and Single-Task Results}
\label{app:rewardfns}

{\setlength{\mathindent}{0cm}\subsection{Reward Functions}
The variables that will be discussed are the following:
\begin{flalign*}
O \in \mathbb{R}^3 &: \text{object position} \\
h \in \mathbb{R}^3 &: \text{hand/gripper position} \\
t \in \mathbb{R}^3 &: \text{target/goal position} \\
h_l \in \mathbb{R}^3 &: \text{position of the left hand/gripper pad} \\
h_r \in \mathbb{R}^3 & : \text{position of the right hand/gripper pad} \\
O_i \in \mathbb{R}^3 &: \text{initial position of the object} \\
h_i \in \mathbb{R}^3 &: \text{initial position of the hand/gripper} \\
g \in \mathbb{R} & : \text{gripper closed/open amount} &&
\end{flalign*}
The following tolerance function is used frequently:

\[   L(x,b_{min},b_{max},m)=\left\{
\begin{array}{ll}
      1 & b_{min}\leq x\leq b_{max} \\
      S\left(\frac{b_{min}-x}{m}, 0.1\right) & x<b_{min} \\
      S\left(\frac{x-b_{max}}{m}, 0.1\right) & x\geq b_{max}
\end{array} \right. \]

Where  S is defined to be a long-tail sigmoid:

\[S(a_1, a_2)=\left(\left(\frac{1}{a_2 - 1} - 1\right)a_1^2 + 1\right)^{-1}\]

With these basics in place, we define a caging tensor that describes behaviour in an axis which intersects the gripper's actuated fingers (in code, the Y axis):

\[C_{LR}(c_1, c_2) = L\left(
\left| \begin{bmatrix} h_{L,(y)} \\ h_{R,(y)} \end{bmatrix} - o_{(y)}\right|,
c_1,
c_2,
\left| \left| \begin{bmatrix} h_{L,(y)} \\ h_{R,(y)} \end{bmatrix} - o_{i,(y)}\right| - c_2\right|
\right)\]

A similar caging value describes behaviour in the other two axes (in code, X and Z axes):

\[C_P(c_3) = L\left(
\lVert o_{(xz)} - h_{(xz)} \rVert,
0,
c_3,
\lVert o_{i,(xz)} - h_{i,(xz)} \rVert - c_3
\right)\]

These get lumped together as follows ($T_{H_0}$ is the Hamacher product):

\[C(c_1, c_2, c_3)=T_{H_0}(T_{H_0}(C_{LR,(0)}, C_{LR,(1)}), C_P(c_3))\]

The caging reward has two modes: medium density and high density. The arguments $c_1, c_2, c_3$ are passed to $C$

\[R_{cage,dense}(c_1, c_2, c_3)=\left\{
\begin{array}{ll}
      0.5(C + T_{H_0}(C, g)) & C > 0.97 \\
      0.5C & otherwise
\end{array} \right. \]

\[R_{cage}(c_1, c_2, c_3, c_4)=\left\{
\begin{array}{ll}
      0.5(L(\lVert o - h \rVert, 0, c_4, \lVert o - h_i \rVert) + T_{H_0}(C, g)) & C > 0.97 \\
      0.5L(\lVert o - h \rVert, 0, c_4, \lVert o - h_i \rVert) & otherwise
\end{array} \right. \]

In each set of expressions given below, the arguments passed to $R_{cage}$ or $R_{cage,dense}$ correspond to $[c_1,c_2,c_3...]$. The caging reward also considers $[h,h_i,o,o_i]$ as described on the previous page, but these arguments are omitted for brevity.

If computation involves a parameter $A$, understand that $A$ is non-zero $iff$ the Sawyer successfully grasps the object. As such, $A$ serves as a post-grasp guidance term.

Common patterns include $A+T_{H_0}(R_{cage}, L(t-o,...))$, $T_{H_0}(1-g, L(o-h,...))$, and $L(t-o,...)+L(o-h,...)$. As a general rule, rewards for simple tasks consist of summed tolerances, while more difficult tasks add complexity in the form of Hamacher Products. The Hamacher Products combine tolerances, grip effort, and/or $R_{cage}$ to produce a smooth, dense reward.

\pagebreak

\subsubsection{Basketball}
\begin{align*}
A=\mathbb{I}_{\lVert o - h \rVert < 0.035 \And g > 0 \And o_{(z)} - o_{i(z)} > 0.01} \cdot
    (1 + L(\lVert \langle1,1,2\rangle \cdot (t - o) \rVert, 0, 0.08, \lVert \langle1,1,2\rangle \cdot (t - o_i) \rVert))
\end{align*}
\[
R=\left\{
  \begin{array}{ll}
\begin{aligned}
  A + T_{H_0}(& R_{cage,dense}(0.025,0.06,0.005), \\
              & L(\lVert \langle1,1,2\rangle \cdot (t - o) \rVert, 0, 0.08, \lVert \langle1,1,2\rangle \cdot (t - o_i) \rVert))
\end{aligned} & \lVert \langle1,1,2\rangle \cdot (t - o) \rVert\geq 0.08  \\
10 & otherwise \\
  \end{array}\right.
\]

\subsubsection{Button Press Top Down}
\[R=\left\{
\begin{array}{ll}
      5T_{H_0}(
        1-g,
        L(\lVert o - h \rVert, 0, 0.01, \lVert o - h_i \rVert))
        & \lVert o - h \rVert > 0.03 \\
      5T_{H_0}(
        1-g,
        L(\lVert o - h \rVert, 0, 0.01, \lVert o - h_i \rVert)) +
        5L(| t_{(z)} - o_{(z)} |, 0, 0.005, | t_{(z)} - o_{i,(z)} |))
        & otherwise
\end{array} \right. \]

\subsubsection{Button Press Top Down Wall}
\[R=\left\{
\begin{array}{ll}
      5T_{H_0}(
        1-g,
        L(\lVert o - h \rVert, 0, 0.01, \lVert o - h_i \rVert))
        & \lVert o - h \rVert > 0.03 \\
      5T_{H_0}(
        1-g,
        L(\lVert o - h \rVert, 0, 0.01, \lVert o - h_i \rVert)) +
        5L(| t_{(z)} - o_{(z)} |, 0, 0.005, | t_{(z)} - o_{i,(z)} |))
        & otherwise
\end{array} \right. \]

\subsubsection{Button Press}
\[R=\left\{
\begin{array}{ll}
      2T_{H_0}(
        g,
        L(\lVert o - h \rVert, 0, 0.05, \lVert o - h_i \rVert))
        & \lVert o - h \rVert > 0.05 \\
      2T_{H_0}(
        g,
        L(\lVert o - h \rVert, 0, 0.05, \lVert o - h_i \rVert)) +
        8L(| t_{(y)} - o_{(y)} |, 0, 0.005, | t_{(y)} - o_{i,(y)} |))
        & otherwise
\end{array} \right. \]

\subsubsection{Button Press Wall}
\[R=\left\{
\begin{array}{ll}
      2T_{H_0}(
        1-g,
        L(\lVert o - h \rVert, 0, 0.01, \lVert o - h_i \rVert))
        & \lVert o - h \rVert > 0.07 \\
      4 + 2g + 4(L(| t_{(y)} - o_{(y)} |, 0, 0.005, | t_{(y)} - o_{i,(y)} |)^2))
        & otherwise
\end{array} \right. \]

\subsubsection{Coffee Button}
\[R=\left\{
\begin{array}{ll}
      2T_{H_0}(
        g,
        L(\lVert o - h \rVert, 0, 0.05, \lVert o - h_i \rVert))
        & \lVert o - h \rVert > 0.05 \\
      2T_{H_0}(
        g,
        L(\lVert o - h \rVert, 0, 0.05, \lVert o - h_i \rVert)) +
        8L(| t_{(y)} - o_{(y)} |, 0, 0.005, | t_{(y)} - o_{i,(y)} |))
        & otherwise
\end{array} \right. \]

\subsubsection{Coffee Pull}
\[
A=\mathbb{I}_{\lVert o - h \rVert < 0.04 \And g > 0} \cdot
    (1 + 5L(\lVert \langle2,2,1\rangle \cdot (t - o) \rVert, 0, 0.05, \lVert \langle2,2,1\rangle \cdot (t - o_i) \rVert))
\]
\[R=\left\{
\begin{array}{ll}
    \begin{aligned}
      A + T_{H_0}(
        & R_{cage}(0.02,0.05,0.05,0.04), \\
        & L(\lVert \langle2,2,1\rangle \cdot (t - o) \rVert, 0, 0.05, \lVert \langle2,2,1\rangle \cdot (t - o_i) \rVert))
    \end{aligned}
        & \lVert \langle2,2,1\rangle \cdot (t - o) \rVert\geq 0.05 \\
      10 & otherwise
\end{array} \right. \]

\subsubsection{Coffee Push}
\[
A=\mathbb{I}_{\lVert o - h \rVert < 0.04 \And g > 0} \cdot
    (1 + 5L(\lVert \langle2,2,1\rangle \cdot (t - o) \rVert, 0, 0.05, \lVert \langle2,2,1\rangle \cdot (t - o_i) \rVert))
\]
\[R=\left\{
\begin{array}{ll}
    \begin{aligned}
      A + T_{H_0}(
        & R_{cage}(0.02,0.05,0.05,0.04), \\
        & L(\lVert \langle2,2,1\rangle \cdot (t - o) \rVert, 0, 0.05, \lVert \langle2,2,1\rangle \cdot (t - o_i) \rVert))
        \end{aligned}
        & \lVert \langle2,2,1\rangle \cdot (t - o) \rVert\geq 0.05 \\
      10 & otherwise
\end{array} \right. \]

\subsubsection{Door Close}
\[R=\left\{
\begin{array}{ll}
    6L(\lVert t - o \rVert, 0, 0.05, \lVert t - o_i \rVert)) +
    3L(\lVert t - h \rVert, 0, 0.012, 0.1 + \lVert h_i - o \rVert)
    & \lVert t - o \rVert\geq 0.05 \\
      10 & otherwise
\end{array} \right. \]

\subsubsection{Door Lock}
\[
R=2T_{H_0}(
    g,
    L(\lVert \langle1,4,2\rangle \cdot (o - h) \rVert, 0, 0.01, \lVert \langle1,4,2\rangle \cdot (o - h_i) \rVert))
+ 8L(|t_{(z)} - o_{i,(z)}|, 0, 0.005, 0.1)
\]

\subsubsection{Door Unlock}
\[R=
\begin{aligned}
2L(&\lVert \langle1,4,2\rangle \cdot (o - h + \langle0, 0.055, 0.07\rangle) \rVert, \\ 
&\;0, \\ 
&\;0.02, \\ 
&\lVert \langle1,4,2\rangle \cdot (o_i - h_i + \langle0, 0.055, 0.07\rangle) \rVert)) + 
8L(|t_{(x)} - o_{i,(x)}|, 0, 0.005, 0.1)
\end{aligned}
\]

\subsubsection{Door Open}
\[
alt = \mathbb{I}_{\lVert h_{(xy)} - o_{(xy)} \lVert > 0.12} \cdot \left(0.4 + 0.04\log\left(\lVert h_{(xy)} - o_{(xy)} \lVert-0.12\right) \right)
\]

\[ready=\left\{
\begin{array}{ll}
T_{H_0}\left(
    L(\lVert h-o-\langle0.05,0.03,-0.01\rangle \rVert, 0, 0.06, 0.5), 
    L(alt-h_{(z)}, 0, 0.01, \frac{alt}{2}),
\right)
& h_{(z)} < alt \\

L(\lVert h-o-\langle0.05,0.03,-0.01\rangle \rVert, 0, 0.06, 0.5)
& otherwise
\end{array} \right. \]

\[R=\left\{
\begin{array}{ll}
2T_{H_0}\left(g, ready\right) +
8\left( 0.2\mathbb{I}_{o_{(\theta)} < 0.03} + 0.8L(o_{(\theta)} + \frac{2\pi}{3}, 0, 0.5, \frac{\pi}{3}) \right)
& |t_{(x)} - o_{(x)}| > 0.08 \\

10 & otherwise
\end{array} \right. \]

\subsubsection{Box Close}
\[
alt = \mathbb{I}_{\lVert h_{(xy)} - o_{(xy)} \lVert > 0.02} \cdot \left(0.4 + 0.04\log\left(\lVert h_{(xy)} - o_{(xy)} \lVert-0.02\right) \right)
\]

\[ready=\left\{
\begin{array}{ll}
T_{H_0}\left(
    L(\lVert h-o \rVert, 0, 0.02, 0.5), 
    L(alt-h_{(z)}, 0, 0.01, \frac{alt}{2}),
\right)
& h_{(z)} < alt \\

L(\lVert h-o \rVert, 0, 0.02, 0.5)
& otherwise
\end{array} \right. \]

\[R=\left\{
\begin{array}{ll}
2T_{H_0}\left(\frac{g+1}{2}, ready\right) +
8\left( 0.2\mathbb{I}_{o_{(z)} > 0.04} + 0.8L(\langle 1, 1, 3 \rangle \lVert t - o \rVert, 0, 0.05, 0.25) \right)
& |t - o| \geq 0.08 \\

10 & otherwise
\end{array} \right. \]

\subsubsection{Drawer Open}
\[
R=5\left(
    L\left(\lVert t - o \rVert, 0, 0.02, 0.2\right) +
    L\left(\lVert (o - h) \cdot \langle3, 3, 1\rangle \rVert, 0, 0.01, \lVert (o_i - h_i) \cdot \langle3, 3, 1\rangle \rVert \right)
    \right)
\]

\subsubsection{Drawer Close}
\[
R=\left\{
    \begin{array}{ll}
        T_{H_0}\left(
            L(\lVert t - o \rVert, 0, 0.05, \lVert t - o_i \rVert - 0.05),
            T_{H_0}\left(
                g,
                L(\lVert o - h \rVert, 0, 0.005, \lVert o_i - h_i \rVert - 0.005)
            \right)
        \right)
            & \lVert t - o \rVert > 0.065 \\
        10
            & otherwise
    \end{array}
\right.
\]

\subsubsection{Faucet Close}

\[
R=\left\{
    \begin{array}{ll}
        4L(\lVert o - h \rVert, 0, 0.01, \lVert o_i - h_i \rVert - 0.01) +
        6L(\lVert t - o \rVert, 0, 0.07, \lVert t - o_i \rVert - 0.07)
            & \lVert t - o \rVert > 0.07 \\
        10
            & otherwise
    \end{array}
\right.
\]

\subsubsection{Faucet Open}

\[
R=\left\{
    \begin{array}{ll}
        \begin{aligned}
         (& 4L(\lVert o - h + \langle-.04,0,.03\rangle \rVert, 0, 0.01, \lVert o_i - h_i \rVert - 0.01)  \\
         & + 6L(\lVert t - o + \langle-.04,0,.03\rangle \rVert, 0, 0.07, \lVert t - o_i \rVert - 0.07) )
         \end{aligned}
        & \lVert t - o + \langle-.04,0,.03\rangle \rVert > 0.07 \\
        
        10
            & otherwise
    \end{array}
\right.
\]

\subsubsection{Hand Insert}

\[
A=\mathbb{I}_{\lVert o - h \rVert < 0.02 \And g > 0} \cdot
    (1 + 7L(\lVert t - o \rVert, 0, 0.05, \lVert t - o_i \rVert))
\]
\[
R=\left\{
    \begin{array}{ll}
        A + 
        T_{H_0}\left(
            R_{cage,dense}(0.015,0.05,0.005),
            L(\lVert t - o \rVert, 0, 0.05, \lVert t - o_i \rVert)
        \right)
            & \lVert t - o \rVert > 0.05 \\
        10
            & otherwise
    \end{array}
\right.
\]

\subsubsection{Pick Place}

\[
A=\mathbb{I}_{\lVert o - h \rVert < 0.02 \And g > 0 \And o_{(z)} > 0.01} \cdot
    (1 + 5L(\lVert t - o \rVert, 0, 0.05, \lVert t - o_i \rVert))
\]
\[
R=\left\{
    \begin{array}{ll}
        A + 
        T_{H_0}\left(
            R_{cage,dense}(0.015,0.05,0.005),
            L(\lVert t - o \rVert, 0, 0.05, \lVert t - o_i \rVert)
        \right)
            & \lVert t - o \rVert > 0.05 \\
        10
            & otherwise
    \end{array}
\right.
\]

\subsubsection{Pick Out Of Hole}
A funnel-shaped surface guides the gripper as it seeks to grab and lift the object; this prevents the gripper from running into the side of the hole in the table. The height (or "altitude") of this surface is given by $alt$ since the variables $h$ and $z$ are already used.
\[
alt = \mathbb{I}_{\lVert h_{(xy)} - o_{i,(xy)} \lVert > 0.03} \cdot
    \left(0.15 + 0.015\log\left(\lVert h_{(xy)} - o_{i,(xy)} \lVert - 0.03\right)\right)
\]
\[
\begin{aligned}
A=&\mathbb{I}_{\lVert o - h \rVert < 0.04 \And g < 0.33 \And o_{(z)} - o_{i,(z)} > 0.02} \\ 
& \bullet
    \left(1 + 5T_{H_0}\left(
        \begin{aligned}
        L(&\lVert t - o \rVert, 0, 0.02, \lVert t - o_i \rVert), \\
         max(&\mathbb{I}_{h_{(z)} > alt}, L(alt - h_{(z)}, 0, 0.01, 0.02))\\
        \end{aligned}
    \right)\right)
\end{aligned}
\]
\[
R=\left\{
    \begin{array}{ll}
        A + 
        T_{H_0}\left(
            R_{cage,dense}(0.015,0.05,0.005),
            L(\lVert t - o \rVert, 0, 0.05, \lVert t - o_i \rVert)
        \right)
            & \lVert t - o \rVert > 0.05 \\
        10
            & otherwise
    \end{array}
\right.
\]

\subsubsection{Plate Slide Back Side}
\[
A=\mathbb{I}_{\lVert o - h \rVert < 0.07 \And h_{(z)} \leq 0.03}
\]
\[
R=\left\{
    \begin{array}{ll}
        A \cdot
        (2 + 7L(\lVert t - o \rVert, 0, 0.05, \lVert t - o_i \rVert)) + 
        (1-A) \cdot
        1.5L(\lVert o - h \rVert, 0, 0.05, \lVert o_i - h_i \rVert - 0.05)
            & \lVert t - o \rVert > 0.05 \\
        10
            & otherwise
    \end{array}
\right.
\]

\subsubsection{Plate Slide Back}
\[
A=\mathbb{I}_{\lVert o - h \rVert < 0.07 \And h_{(z)} \leq 0.03}
\]
\[
R=\left\{
    \begin{array}{ll}
        A \cdot
        (2 + 7L(\lVert t - o \rVert, 0, 0.05, \lVert t - o_i \rVert)) + 
        (1-A) \cdot
        1.5L(\lVert o - h \rVert, 0, 0.05, \lVert o_i - h_i \rVert - 0.05)
            & \lVert t - o \rVert > 0.05 \\
        10
            & otherwise
    \end{array}
\right.
\]

\subsubsection{Plate Slide Side}
\[
A=\mathbb{I}_{\lVert o - h \rVert < 0.07 \And h_{(z)} \leq 0.03}
\]
\[
R=\left\{
    \begin{array}{ll}
        A \cdot
        (2 + 7L(\lVert t - o \rVert, 0, 0.05, \lVert t - o_i \rVert)) + 
        (1-A) \cdot
        1.5L(\lVert o - h \rVert, 0, 0.05, \lVert o_i - h_i \rVert - 0.05)
            & \lVert t - o \rVert > 0.05 \\
        10
            & otherwise
    \end{array}
\right.
\]

\subsubsection{Plate Slide}
\[
R=\left\{
    \begin{array}{ll}
    8T_{H_0}(
        L(\lVert t - o \rVert, 0, 0.05, \lVert t - o_i \rVert),
        L(\lVert o - h \rVert, 0, 0.05, \lVert o_i - h_i \rVert)
    ) & \lVert t - o \rVert \geq 0.05 \\
    10 & otherwise
    \end{array}
\right.
\]

\subsubsection{Handle Press Side}
\[
R=\left\{
    \begin{array}{ll}
    10T_{H_0}(
        L(| t_{(z)} - o_{(z)} |, 0, 0.05, | t_{(z)} - o_{i,(z)} |),
        L(\lVert o - h \rVert, 0, 0.02, \lVert o_i - h_i \rVert - 0.02)
    ) & \lVert t - o \rVert > 0.05 \\
    10 & otherwise
    \end{array}
\right.
\]

\subsubsection{Handle Press}
\[
R=\left\{
    \begin{array}{ll}
    10T_{H_0}(
        L(| t_{(z)} - o_{(z)} |, 0, 0.05, | t_{(z)} - o_{i,(z)} |),
        L(\lVert o - h \rVert, 0, 0.02, \lVert o_i - h_i \rVert - 0.02)
    ) & \lVert t - o \rVert > 0.05 \\
    10 & otherwise
    \end{array}
\right.
\]

\subsubsection{Handle Pull}
\[
A=\mathbb{I}_{\lVert o - h \rVert < 0.035 \And g > 0 \And o_{(z)} - o_{i,(z)} > 0.01} \cdot
    (1 + 5L(| t_{(z)} - o_{(z)} |, 0, 0.05, | t_{(z)} - o_{i,(z)} |))
\]
\[
R=\left\{
    \begin{array}{ll}
        A + 
        T_{H_0}\left(
            R_{cage,dense}(0.022,0.05,0.01),
            L(| t_{(z)} - o_{(z)} |, 0, 0.05, | t_{(z)} - o_{i,(z)} |)
        \right)
            & \lVert t - o \rVert > 0.05 \\
        10
            & otherwise
    \end{array}
\right.
\]

\subsubsection{Handle Pull Side}
\[
A=\mathbb{I}_{\lVert o - h \rVert < 0.035 \And g > 0 \And o_{(z)} - o_{i,(z)} > 0.01} \cdot
    (1 + 5L(| t_{(z)} - o_{(z)} |, 0, 0.05, | t_{(z)} - o_{i,(z)} |))
\]
\[
R=\left\{
    \begin{array}{ll}
        A + 
        T_{H_0}\left(
            R_{cage,dense}(0.032,0.06,0.01),
            L(| t_{(z)} - o_{(z)} |, 0, 0.05, | t_{(z)} - o_{i,(z)} |)
        \right)
            & \lVert t - o \rVert > 0.05 \\
        10
            & otherwise
    \end{array}
\right.
\]

\subsubsection{Reach}
\[
R=10L(\lVert t - h \rVert, 0, 0.05, \lVert t - h_i \rVert)
\]

\subsubsection{Reach Wall}
\[
R=10L(\lVert t - h \rVert, 0, 0.05, \lVert t - h_i \rVert)
\]

\subsubsection{Push}
\[
A=\mathbb{I}_{\lVert o - h \rVert < 0.02 \And g > 0}
\]
\[
R=\left\{
    \begin{array}{ll}
    (A+1) \cdot R_{cage,dense}(0.015,0.05,0.005) +
    A \cdot (1 + 5L(\lVert t - o \rVert, 0, 0.05, \lVert t - o_i \rVert))
        & \lVert t - o \rVert > 0.05 \\
    10
        & otherwise
    \end{array}
\right.
\]

\subsubsection{Sweep Into Goal}
Note: This technically uses a $R_{cage,dense}$ function with slightly different margin parameters than the one described above (they are constant rather than dynamic), but the behaviour is mostly the same.
\[
R=\left\{
    \begin{array}{ll}
    \begin{aligned}
        & (2R_{cage,dense}(0.02,0.05,0.01) \\
        & + 2T_{H_0}\left(
            R_{cage,dense}(0.02,0.05,0.01),
            L(\lVert t - o \rVert, 0, 0.05, \lVert t - o_i \rVert)
        \right))
    \end{aligned}
            & \lVert t - o \rVert > 0.05 \\
        10
            & otherwise
    \end{array}
\right.
\]

\subsubsection{Sweep}
Note: This technically uses a $R_{cage,dense}$ function with slightly different margin parameters than the one described above (they are constant rather than dynamic), but the behaviour is mostly the same.
\[
R=\left\{
    \begin{array}{ll}
        \begin{aligned}
        & (2R_{cage,dense}(0.02,0.05,0.01) \\
        & + 2T_{H_0}\left(
            R_{cage,dense}(0.02,0.05,0.01),
            L(\lVert t - o \rVert, 0, 0.05, \lVert t - o_i \rVert)
        \right))
        \end{aligned}
            & \lVert t - o \rVert > 0.05 \\
        10
            & otherwise
    \end{array}
\right.
\]

\subsubsection{Push Back}
Note: This technically uses a $R_{cage,dense}$ function with slightly different margin parameters than the one described above (they are constant rather than dynamic), but the behaviour is mostly the same.
\[
A=\mathbb{I}_{\lVert o - h \rVert < 0.01 \And 0 < g < 0.55 \And \lVert t - o_i \rVert - \lVert t - o \rVert > 0.01} \cdot
    (1 + 5L(\lVert t - o \rVert, 0, 0.05, \lVert t - o_i \rVert))
\]
\[
R=\left\{
    \begin{array}{ll}
        A + 
        T_{H_0}\left(
            R_{cage,dense}(0.01,0.05,0.01),
            L(\lVert t - o \rVert, 0, 0.05, \lVert t - o_i \rVert)
        \right)
            & \lVert t - o \rVert > 0.05 \\
        10
            & otherwise
    \end{array}
\right.
\]

\subsubsection{Window Open}
\[
R=10T_{H_0}(
    L(| t_{(x)} - o_{(x)} |, 0, 0.05, | t_{(x)} - o_{i,(x)} |),
    L(\lVert o - h \rVert, 0, 0.02, \lVert o_i - h_i \rVert - 0.02)
)
\]

\subsubsection{Window Close}
\[
R=10T_{H_0}(
    L(| t_{(x)} - o_{(x)} |, 0, 0.05, | t_{(x)} - o_{i,(x)} |),
    L(\lVert o - h \rVert, 0, 0.02, \lVert o_i - h_i \rVert - 0.02)
)
\]

\subsubsection{Dial Turn}
\[
\begin{aligned}
R=10T_{H_0}(
    & L(\lVert t - o \rVert, 0, 0.05, \lVert t - o_i \rVert - 0.05) , \\
    & \begin{aligned}
        T_{H_0}(& g, \\
        L(&\lVert o - h + \langle0.05,0.02,0.09\rangle \rVert , \\ 
        & 0, \\
        &0.005, \\
        &\lVert o_i - h_i + \langle0.05,0.02,0.09\rangle \rVert - 0.005)))
    \end{aligned}
\end{aligned}
\]

\subsubsection{Bin Picking}
Two funnel-shaped surfaces guide the gripper as it seeks to carry the object between the two bins; this prevents the gripper from running into the side of the bins. The height (or "altitude") of this surface is given by $alt$ since the variables $h$ and $z$ are already used.
\[
\begin{aligned}
alt = min (&\mathbb{I}_{\lVert h_{(xy)} - o_{i,(xy)} \lVert > 0.03} \cdot
    \left(0.2 + 0.02\log \left(\lVert h_{(xy)} - o_{i,(xy)} \lVert - 0.03\right)\right), \\
    & \mathbb{I}_{\lVert h_{(xy)} - t_{(xy)} \lVert > 0.03} \cdot
    \left(0.2 + 0.02\log\left(\lVert h_{(xy)} - t_{(xy)} \lVert - 0.03\right)\right))
\end{aligned}
\]
\[
A=\mathbb{I}_{\lVert o - h \rVert < 0.04 \And g < 0.43 \And o_{(z)} - o_{i,(z)} > 0.02} \cdot
    \left(1 + 5T_{H_0}\left(
        \begin{aligned}
        &L(\lVert t - o \rVert, 0, 0.05, \lVert t - o_i \rVert), \\
        & max(\mathbb{I}_{h_{(z)} > alt}, L(alt - h_{(z)}, 0, 0.01, 0.05))\\
        \end{aligned}
    \right)\right)
\]
\[
R=\left\{
    \begin{array}{ll}
        A + 
        T_{H_0}\left(
            R_{cage,dense}(0.015,0.05,0.01),
            L(\lVert t - o \rVert, 0, 0.05, \lVert t - o_i \rVert)
        \right)
            & \lVert t - o \rVert > 0.05 \\
        10
            & otherwise
    \end{array}
\right.
\]

\subsubsection{Assembly}
In addition to the components described below, the assembly reward is weighted by how level the object is (tilted object quaternions are penalized).

\[
alt = \mathbb{I}_{\lVert t_{(xy)} - o_{(xy)} \lVert > 0.02} \cdot \left(0.4 + 0.04\log\left(\lVert t_{(xy)} - o_{(xy)} \lVert-0.02\right) \right)
\]

\[
A=0.1\mathbb{I}_{o_{(z)} > 0.02 or \lVert t_{(xy)} - o_{(xy)} \lVert < 0.02} + 
    0.9L(\langle 1, 1, 3 \rangle \langle t_{(x)} - o_{(x)}, t_{(y)} - o_{(y)}, alt - o_{(z)} \rangle, 0, 0.02, 0.4)
\]

\[R=\left\{
\begin{array}{ll}
2R_{cage,dense}(0.015, 0.02, 0.01) + 8A
& |t_{(x)} - o_{(x)}| > 0.02 \\

10 & otherwise
\end{array} \right. 
\]

\subsubsection{Disassemble}
In addition to the components described below, the disassemble reward is weighted by how level the object is (tilted object quaternions are penalized).

\[R=\left\{
\begin{array}{ll}
2R_{cage,dense}(0.015, 0.02, 0.01) +
6 \left( 0.1\mathbb{I}_{o_{(z)} > 0.02} + 0.9L(\lVert t - o \rVert, 0, 0.02, 0.2) \right)
& o_{(z)} > t_{(z)} \\

10 & otherwise
\end{array} \right. 
\]

\subsubsection{Hammer}
In addition to the components described below, the hammer reward is weighted by how level the object is (tilted object quaternions are penalized).

\[R=\left\{
\begin{array}{ll}
2R_{cage,dense}(0.015, 0.02, 0.01) +
6 \left( 0.1\mathbb{I}_{o_{(z)} > 0.02} + 0.9L(\lVert t - o \rVert, 0, 0.02, 0.2) \right)
& |o_{(y)} - o_{i,(y)}| > 0.09 \\

10 & otherwise
\end{array} \right. 
\]

\subsubsection{Lever Pull}

\[
\begin{aligned}
R=10T_{H_0}(
    &L\left(\lVert t - o \rVert, 0, 0.04, \lVert t - o_i \rVert\right), \\
    &L\left(
        \langle4,1,4\rangle \cdot (h - o + \langle0,0.055,0.07\rangle),
        0,
        0.02,
        \langle4,1,4\rangle \cdot (h_i - o_i + \langle0,0.055,0.07\rangle)
    \right)
)\end{aligned}
\]

\subsubsection{Stick Push}
Note: \textit{a} is the second object in the environment, which in this case is a thermos.
\[
R=\left\{
\begin{array}{ll}
\begin{aligned}
& 2 + 5L(\lVert t-o \rVert, 0, 0.12, \lVert t-o_i \rVert - 0.12) \\
& + 3L(\lVert t-a \rVert, 0, 0.12, \lVert t-a_i \rVert - 0.12)\\
\end{aligned}
& \lVert h-o \rVert < 0.02, g>0, o_{(z)}-o_{i,(z)}>0.01, \lVert t-a \rVert > 0.12 \\

10
& \lVert h-o \rVert < 0.02, g>0, o_{(z)}-o_{i,(z)}>0.01, \lVert t-a \rVert \leq 0.12 \\

R_{cage,dense}(0.04,0.05,0.01)
& otherwise
\end{array} \right. \]

\subsubsection{Stick Pull}
Note: \textit{a} is the second object in the environment, which in this case is a thermos.
\textit{in} is a condition involving lots of vector offsets from the object observations. It indicates whether the stick is inserted into the thermos' handle or not. The variable \textit{stick\_in\_place}, and \textit{stick\_grabbed} have also been defined so that the reward function fits on one page.

$\textit{stick\_in\_place} = L(\lVert (o-a) \cdot \langle 1,1,2 \rangle \rVert, 0, 0.12, \lVert (o_i-a_i) \cdot \langle 1,1,2 \rangle \rVert)$

$\textit{stick\_grabbed} = \lVert h-o \rVert < 0.02, g>0, o_{(z)}-o_{i,(z)}>0.01$

\begin{align*}
R=\left\{\begin{array}{ll} 1 + 6 \cdot stick\_in\_place &  stick\_grabbed, \lnot in, \lVert t-a \rVert > 0.12 \\
\begin{aligned}
& 6
+ stick\_in\_place
+ 2L(\lVert t-o \rVert, 0, 0.12, \lVert t-o_i \rVert) \\ 
& + L(\lVert t-a \rVert, 0, 0.12, \lVert t-a_i \rVert)
\end{aligned}
& stick\_grabbed, in, \lVert t-a \rVert > 0.12 \\
10
& stick\_grabbed, in, \lVert t-a \rVert \leq 0.12 \\
T_{H_0}(R_{cage,dense}(0.014,0.05,0.01), stick\_in\_place)
& otherwise
\end{array} \right.
\end{align*}

\subsubsection{Shelf Place}
In addition to the components described below, the shelf-place reward includes negative components that help avoid collision with the shelf.
\[
A=\mathbb{I}_{\lVert o - h \rVert < 0.025 \And g > 0 \And o_{(z)}>0.01} \cdot
    (1 + 5L(\lVert t - o \rVert, 0, 0.05, \lVert t - o_i \rVert))
\]
\[R=\left\{
\begin{array}{ll}
      A + T_{H_0}(
        R_{cage}(0.01,0.02,0.05,0.01),
        L(\lVert t - o \rVert, 0, 0.05, \lVert t - o_i \rVert))
        & \lVert t - o \rVert\geq 0.05 \\
      10 & otherwise
\end{array} \right. \]

\subsubsection{Peg Insert}
In addition to the components described below, the peg-insert reward includes negative components that help avoid collision with the hole/box into which the peg gets inserted.
\[
A=\mathbb{I}_{\lVert o - h \rVert < 0.08 \And g > 0 \And o_{(z)}>0.01} \cdot
    (1 + 5L(\lVert \langle2,2,1\rangle \cdot (t - o) \rVert, 0, 0.05, \lVert \langle2,2,1\rangle \cdot (t - o_i) \rVert))
\]
\[R=\left\{
\begin{array}{ll}
      \begin{aligned}
      A + T_{H_0}(&R_{cage}(0.0075,0.01,0.03,0.005),\\
      
        &L(\lVert \langle2,2,1\rangle \cdot (t - o) \rVert, 0, 0.05, \lVert \langle2,2,1\rangle \cdot (t - o_i) \rVert))
        \end{aligned}
        & \lVert \langle2,2,1\rangle \cdot (t - o) \rVert\geq 0.07 \\
      10 & otherwise
      
\end{array} \right. \]

\subsubsection{Peg Unplug}
\[
A=\mathbb{I}_{\lVert o - h \rVert < 0.035 \And g > 0.5 \And o_{(x)}-o_{i,(x)}>0.015} \cdot
    (1 + 5L(\lVert t - o \rVert, 0, 0.05, \lVert t - o_i \rVert))
\]
\[R=\left\{
\begin{array}{ll}
      A + 2R_{cage}(0.01,0.025,0.05,0.005)
        & \lVert t - o \rVert\geq 0.07 \\
      10 & otherwise
\end{array} \right. \]

\subsubsection{Soccer}
In addition to the components described below, the soccer reward function includes parameters to fine-tune movements near the goal line.
\[R=\left\{
\begin{array}{ll}
      \begin{aligned}
      &3R_{cage}(0.013,0.023,0.05,0.005) \\
      &+ 6.5L(\lVert \langle3,1,1\rangle(t - o) \rVert, 0, 0.07, \lVert
      \langle3,1,1\rangle(t - o_i) \rVert)
      \end{aligned}
        & \lVert \langle3,1,1\rangle(t - o) \rVert\geq 0.07 \\
      10 & otherwise
\end{array} \right. \]

\subsubsection{Pick Place Wall}
The pick-place-wall reward is essentially two pick-place rewards stacked on top of one another. The first pick-place reward incentivizes movement to a neutral midpoint above the wall (to avoid running into it). The second pick-place reward incentivizes movement to the target position. The math is such that there is no discontinuity between the two reward components.

\subsubsection{Push Wall}
The push-wall reward is the same as the pick-place-wall reward, but without incentives to pick up the object. Additionally, the midpoint is configured to be next to the wall (so that policies push the object around the wall) rather than above the wall.
}

\begin{table}[h]
    \centering
    \begin{tabular}{lc}
\toprule
\footnotesize Task & Success Metric  \\
\midrule
faucet-open &  $\1_{\|o - t\|_2 < 0.07}$\\
sweep & $\1_{\|o - t\|_2 < 0.05}$\\
pick-out-of-hole & $\1_{\|o - t\|_2 < 0.07}$\\
faucet-close & $\1_{\|o - t\|_2 < 0.07}$\\
push & $\1_{\|o - t\|_2 < 0.05}$\\
stick-push & $\1_{\|o - t\|_2 < 0.12} \;and\; \1_{grasped(o)}$\\
coffee-button & $\1_{\|o - t\|_2 < 0.02}$\\
handle-pull-side & $\1_{\|o - t\|_2 < 0.08}$\\
basketball & $\1_{\|o - t\|_2 < 0.08}$\\
stick-pull & $\1_{\|o - t\|_2 < 0.12} \;and\; \1_{grasped(o)}$\\
sweep-into & $\1_{\|o - t\|_2 < 0.05}$\\
disassemble & $\1_{o_z - o_z initial \;>\; 0.15}$\\
assembly & $\1_{\|o - t\|_2 < 0.02} \;and\; \1_{g_z-o_z > 0}$\\
shelf-place & $\1_{\|o - t\|_2 < 0.07}$\\
coffee-push & $\1_{\|o - t\|_2 < 0.07}$\\
handle-press-side & $\1_{\|o - t\|_2 < 0.02}$\\
hammer & $\1_{nail \;travels\; >\ 0.09 \;into \;wood \;block}$\\
plate-slide & $\1_{\|o - t\|_2 < 0.07}$\\
plate-slide-side & $\1_{\|o - t\|_2 < 0.07}$\\
button-press-wall & $\1_{\|o - t\|_2 < 0.03}$\\
handle-press & $\1_{\|o - t\|_2 < 0.02}$\\
handle-pull & $\1_{\|o - t\|_2 < 0.05}$\\
soccer & $\1_{\|o - t\|_2 < 0.07}$\\
plate-slide-back-side & $\1_{\|o - t\|_2 < 0.07}$\\
plate-slide-back & $\1_{\|o - t\|_2 < 0.07}$\\
drawer-close & $\1_{\|o - t\|_2 < 0.055}$\\
reach & $\1_{\|o - t\|_2 < 0.05}$\\
button-press-topdown-wall & $\1_{\|o - t\|_2 < 0.02}$\\
reach-wall & $\1_{\|o - t\|_2 < 0.05}$\\
peg-insert-side & $\1_{\|o - t\|_2 < 0.07}$\\
push-wall & $\1_{\|o - t\|_2 < 0.07}$\\
pick-place-wall & $\1_{\|o - t\|_2 < 0.07}$\\
button-press & $\1_{\|o - t\|_2 < 0.02}$\\
button-press-topdown & $\1_{\|o - t\|_2 < 0.02}$\\
pick-place & $\1_{\|o - t\|_2 < 0.07}$\\
push-back & $\1_{\|o - t\|_2 < 0.07}$\\
coffee-pull & $\1_{\|o - t\|_2 < 0.07}$\\
peg-unplug-side & $\1_{\|o - t\|_2 < 0.07}$\\
dial-turn & $\1_{\|o - t\|_2 < 0.07}$\\
lever-pull & $\1_{rad(o) - rad(t) < \pi / 24}$\\
window-close & $\1_{\|o - t\|_2 < 0.05}$\\
window-open & $\1_{\|o - t\|_2 < 0.05}$\\
door-open & $\1_{\|o - t\|_2 < 0.08}$\\
door-close & $\1_{\|o - t\|_2 < 0.08}$\\
drawer-open & $\1_{\|o - t\|_2 < 0.03}$\\
hand-insert & $\1_{\|o - t\|_2 < 0.05}$\\
box-close & $\1_{\|o - t\|_2 < 0.08}$\\
door-lock & $\1_{\|o - t\|_2 < 0.02}$\\
door-unlock & $\1_{\|o - t\|_2 < 0.02}$\\
bin-picking & $\1_{\|o - t\|_2 < 0.05}$\\
\bottomrule
\end{tabular}
\vspace{0.2cm}
    \caption{A list of success metrics used for each of the Meta-World tasks. All units are in meters.}
    \label{tbl:task_metrics}
\end{table}

\end{document}